\documentclass[graybox]{svmult}

\usepackage{type1cm}        

\usepackage{makeidx}         
\usepackage{graphicx}        
\usepackage{multicol}        
\usepackage[bottom]{footmisc}
\usepackage{newtxtext}       %
\usepackage{newtxmath}       
\usepackage{natbib}

\usepackage{fancyhdr}
\fancypagestyle{arxivhdr}
{
   \fancyhf{}
   \setlength{\headheight}{40pt} 
\fancyfoot[C]{This chapter appears in: Ang, M.H., Khatib, O., Siciliano, B. (eds) Encyclopedia of Robotics. Springer, Berlin, Heidelberg}
\fancyhead[C]{\footnotesize Please cite this chapter as:\\
Andreasson, H., Grisetti, G., Stoyanov, T., Pretto, A. (2023). Sensors for Mobile Robots. In: Ang, M.H., Khatib, O., Siciliano, B. (eds) Encyclopedia of Robotics. Springer, Berlin, Heidelberg. \url{https://doi.org/10.1007/978-3-642-41610-1\_159-1}}
}

\makeindex             

\def\secref#1{Sec.~\ref{#1}}
\def\figref#1{Fig.~\ref{#1}}
\def\tabref#1{Tab.~\ref{#1}}
\def\eqref#1{Eq.~(\ref{#1})}

\begin{document}
\title*{Sensors for Mobile Robots}
\author{Henrik Andreasson, Giorgio Grisetti, Todor Stoyanov, and Alberto Pretto}
\institute{Andreasson and Stoyanov \at School of Science and Technology, \"{O}rebro University, S-701 82 \"{O}rebro, Sweden, \email{\{henrik.andreasson, todor.stoyanov\}@oru.se}
\and Grisetti \at Department of Computer, Control, and Management Engineering ``Antonio Ruberti``, Sapienza University of Rome, Italy \email{grisetti@diag.uniroma1.it}
\and Pretto \at Department of Information Engineering, University of Padova, Italy \email{alberto.pretto@dei.unipd.it}}
%
%

\maketitle
\thispagestyle{arxivhdr}
\vspace{-2.0cm}
\section*{Definition}
A sensor is a device that converts a physical parameter or an
environmental characteristic (e.g., temperature, distance, speed,
etc.) into a signal that can be digitally measured and processed
to perform specific tasks.
Mobile robots need sensors to measure properties of their environment,
thus allowing for safe navigation, complex perception and corresponding actions, and effective interactions with other agents that populate it.

\section{Overview}

Sensors used by mobile robots range from simple tactile sensors, such as bumpers, 
to complex vision-based sensors such as structured light RGB-D cameras. All of them provide a
digital output (e.g., a string, a set of values, a matrix, etc.) that
can be processed by the robot's computer. Such output is typically obtained by discretizing
one or more analog electrical signals by using an Analog to Digital Converter (ADC)
included in the sensor.\\

In this chapter we present the most common sensors used in mobile robotics,
providing an introduction to their taxonomy, basic features, and specifications.
The description of the functionalities and the types of applications follows a bottom-up approach: 
the basic principles and components on which the sensors are based are presented 
before describing real-world sensors (starting from \secref{sec:proximity}), which are generally based on multiple technologies and basic devices. A sensor can be categorized as an input type \textit{transducer}.
A transducer is a device that converts a signal from one form of energy 
(e.g., mechanical, thermal, etc.) into another. Transducers can be used to either 
inject energy in the environment (\textit{emitters}, or \textit{actuators})
or to capture the amount of energy from the environment (\textit{receivers}).  
A receiver that converts a measurable quantity, such as light or sound, 
to an electrical signal is called a \textit{sensor}. 
In mobile robotics, it is common to define as a sensor also an ensemble
of transducers and other devices, packaged together, that cooperate for a specific sensing function.
For example, ranging sensors based on sound (sonar or ultrasonic sensors) are composed of both a receiver and
an actuator synchronized together to measure distances. Following this definition, transducers
are building blocks of sensors: a sensor is composed of one or more receivers, 
zero or more cooperating actuators, and/or other devices (mechanical, optical, etc.)
needed to observe the measured phenomena. 

\section{Sensors Classifications}

Sensors can be classified in several ways; here the most common classifications used in
robotics are reported, based on the source of the excitation signal, the measurement domain,
and measurement type.\\

\textbf{Excitation Signal Source (Passive and Active Sensors)}: A
sensor is generally classified as \textit{passive} if it does not
require, except for the signal amplification and digital conversion,
a power supply to operate.  A passive sensor changes its output in
response to an excitation signal generated by an external phenomenon,
e.g., a microphone that senses a human voice.  On the other side, an
\textit{active} sensor requires an external power supply to self-generate 
the excitation signal, and it measures the environmental
reaction to such a signal, e.g. a sonar that emits a sound pulse and
then senses the reflected wave of the pulse. An alternative definition
used in mobile robotics indicates passive sensors as sensors that
include only receiver transducers, and active sensors as sensors that
also include one or more emitter transducers.\\

\textbf{Measurement Domain (Proprioceptive and Exteroceptive Sensors)}: 
A sensor is classified as \textit{proprioceptive} if it measures a quantity that
depends only on the internal robot system and its current internal state
(e.g. wheels position, rotational speed, etc.). On the other side, an \textit{exteroceptive} 
sensor measures a quantity that depends on both the robot state (e.g., its position) and the
environment surrounding the robot, for example, the distance to the closest obstacle.\\

\textbf{Measurement Type}:
A sensor can be classified on the basis of the type of measurement performed
(e.g., robot speed, global position), as reported in \tabref{tab:meas_types}, which also includes
some sample sensors from which these measurements can be obtained. It is noteworthy to highlight
that from some sensors it is possible to get or derive more than one type of measurement, e.g. 
from the robot motors encoders, it is possible to derive both the robot position and the robot speed.
Likewise, several complex sensors used in mobile robotics are actually composed of different basic sensors
embedded in the same case, to obtain measurements of several types at the same time, e.g. an 
RGB-D camera captures both the three-dimensional (3D) structure and an image of the perceived 
scene, so it is a sensor able to measure both the range and the light intensity.\\

\begin{table}
\caption{Measurement type classification of the sensors used in mobile robots}
\label{tab:meas_types}
\scriptsize
\begin{tabular}{p{3.4cm}p{4cm}p{4cm}}
\hline\noalign{\smallskip}
Measurement Type & Example Sensors& Example Applications\\
\noalign{\smallskip}\svhline\noalign{\smallskip}
Physical contact & Bumpers, Contact switches & Object manipulation\\\\
Proximity & Reflective photocells & Emergency stop functions\\\\
Acceleration and velocity& Accelerometers, Gyroscopes, Wheels encoders & Dead Reckoning (i.e., short term position estimation)\\\\
Relative position and orientation & Wheels encoders, Gyroscopes & Indoor robot localization and mapping\\\\
Global position and orientation & Global Navigation Satellite System (GNSS), Magnetometers & Outdoor robot localization and mapping\\\\
Range & LiDAR, Sonar, Radars, Stereo cameras, time-of-flight cameras & Object localization, Obstacle avoidance, localization and mapping\\\\
Light intensity and color & RGB cameras & Place recognition (e.g., for loop closure detection)\\\\
\noalign{\smallskip}\hline\noalign{\smallskip}
\end{tabular}
\vspace*{-12pt}
\end{table}

\section{Sensors Characterization and Specifications}\label{sec:sensor_specs}

Manufacturers provide the specifications of their sensors as a set of metrics 
designed to characterize and describe the operating mode and to
measure the sensor performances; the most common are listed below.\\

\textbf{Linearity and Non-linearity}: A sensor is defined linear if its response $y$ to 
the measured stimulus $x$ is represented by a linear or affine function, e.g., in the one-dimensional case $y (x) = \alpha x + \beta$, $\alpha, \beta \in \mathbb{R}$. 
Linearity plays a major role in interpreting the signal. 
Highly non-linear sensors are usually more complex to model,
and the quantization noise tends to vary with the magnitude of the measurement.\\


\textbf{Measurement Range and Dynamic Range}:
The measurement range $[x_{min},x_{max}]$ is represented by the smallest and largest values of the 
sensed signal that can be measured by the sensor. Stimulus outside 
such interval can not be sensed, or they are measured with unacceptable errors, 
or they can damage the sensor. The ratio between $x_{max}$ and $x_{min}$ is called 
\textit{dynamic range}, often represented by its base-10 logarithm multiplied by 20
and measured in decibels.\\

\textbf{Sensitivity}:
The sensitivity of a sensor is defined as the slope $dy/dx $ of the sensor response $y$.
It is a measurement of how much the output of a sensor changes as a result of a change 
in the measured quantity. An effective sensor should have high sensitivity and, for 
linear sensors, it is a constant value. A sensor that, for some input signal changes,
does not respond with any output changes is in a \textit{saturation} state. 
This often happens to a sensor that is working outside its \textit{measurement range}.\\

\begin{figure}[ht!]
   \centering
   \includegraphics[width=\columnwidth]{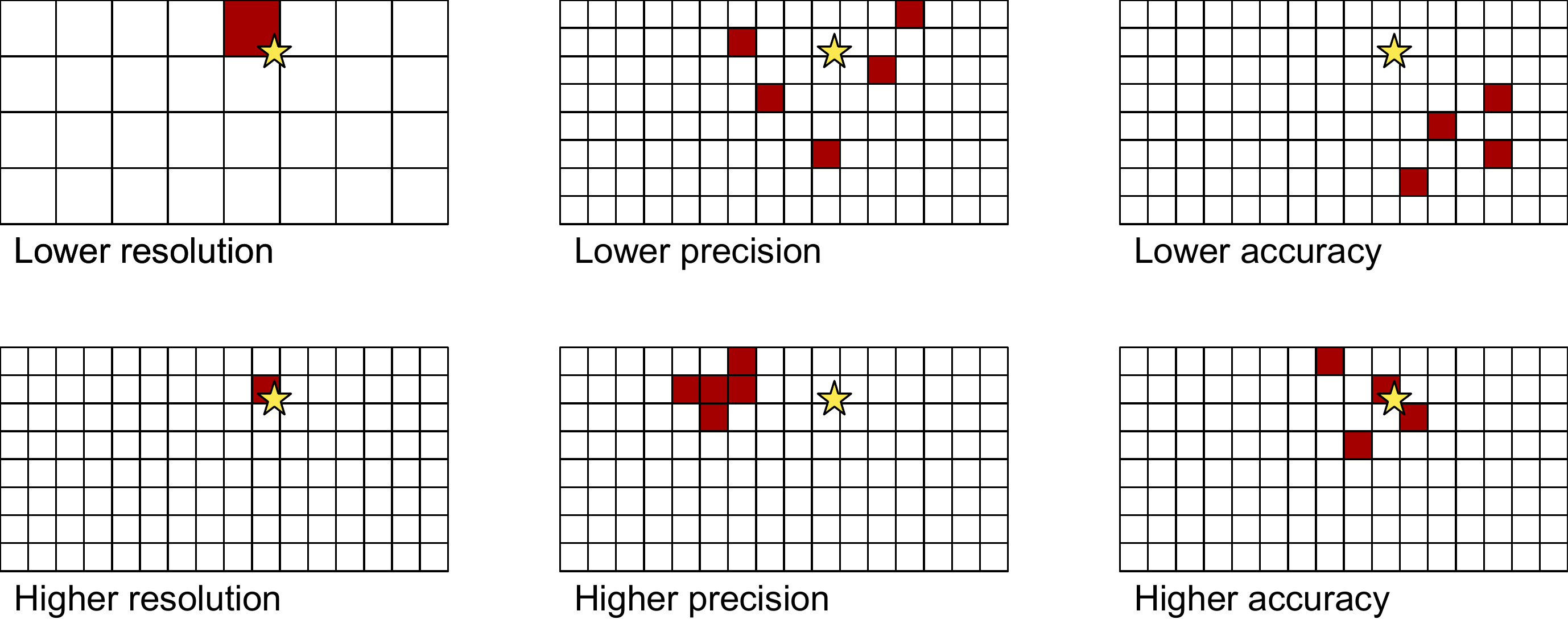}
   \caption{A visual representation of the resolution, precision, and accuracy parameters 
   for a sensor that provides a two-dimensional (2D) output position, e.g., the position of the robot inside
   a planar environment. The yellow star is the ideal, perfect measurement (i.e., 
   the ground truth robot position) while the red squares represent a set of output measurements
   for the steady position represented by the yellow star.}
   \label{fig:res_prec_acc}
\end{figure}
    
\textbf{Resolution}:
The resolution is the minimum variation of the measured signal 
that can produce a detectable change in the sensor output. For a linear,
one-dimensional digital sensor, the resolution can be typically evaluated as the ratio
$(x_{max}-x_{min})/\#_{y}$, where $\#_{y}$ is the number of possible 
discrete output values provided by the sensor. In \figref{fig:res_prec_acc} (left) is reported a
representation of two possible resolutions for a sensor that provides a two-dimensional 
output.\\

\textbf{Precision}:
Precision is a statistical parameter that describes the \textit{reproducibility} of the 
sensor measurements given a steady sensed signal. For such a signal, an ideal sensor 
with infinite precision should provide the same measured output over time.
Real sensors instead provide a range of values over time, 
statistically distributed with respect to some probability density function (\figref{fig:res_prec_acc}, center).
Typically, the precision is evaluated by assuming this density to be Gaussian,
so it can be evaluated by computing the variance of a set of sensor readings
for a steady sensed signal.\\

\textbf{Accuracy}:
The accuracy quantifies the \textit{correctness} of  output provided by a sensor compared with 
the real value of the measured signal (\figref{fig:res_prec_acc}, right). The accuracy can be assessed by taking the 
difference between the average of the measurements of a steady sensed signal 
and its true value; an estimate of the true value can be measured for instance by using 
a sensor with a superior accuracy, or through precise experimental design.\\

\textbf{Bandwidth}:
The sensor bandwidth (typically, represented by a maximum frequency)
quantifies how it behaves for inputs that evolve with different frequencies. A sensor with
low bandwidth can't properly measure high-frequency changes in the measured input 
(e.g, vibrations, motions, etc.); on the other hand, the bandwidth is often limited by a
low-pass filter to avoid to measure high-frequency noise components.\\

\textbf{Response Time}:
The response time is the duration of the period of time that elapses between a change
in the input measured signal and when the sensor output changes accordingly.\\

\section{Basic Sensor Components}


Sensors that are typically installed on mobile robots are composed of a set of basic devices 
(receivers and actuators) that directly measure or generate a physical 
quantity. These components are often integrated and 
packaged in a single chip to form the so-called Microelectromechanical 
Systems (MEMS). In this section, the most common forms of basic receivers 
and actuators used to build more complex sensors are briefly addressed, 
classified according to the basic physical quantity that they measure 
or generate.

\subsection{Force and Deformation}\label{sec:basic_force}
\begin{figure}[ht!]
   \centering
   \includegraphics[width=\columnwidth]{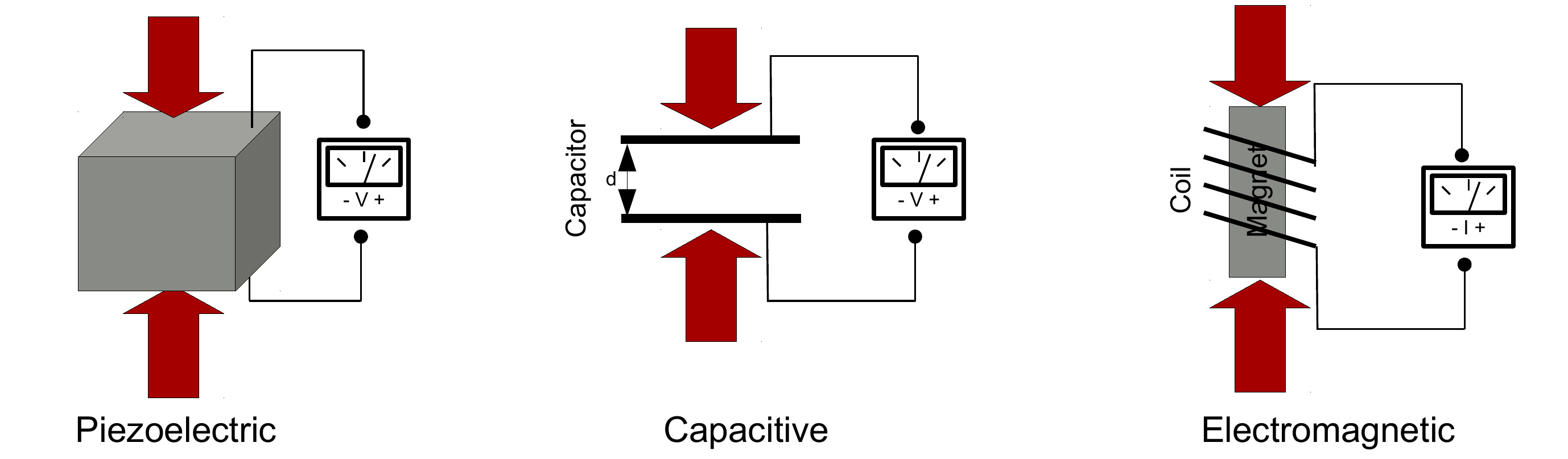}
   \caption{Overview of basic force transducers.}
   \label{fig:force_sensors}
\end{figure}

Force transducers operate by measuring or imposing a deformation to a body
subject to that force. This section focuses only on elementary devices
capable to exert/sense elementary linear deformations (\figref{fig:force_sensors}); 
electrical motors or dynamos that are complex mechanical compounds are not discussed.

\textbf{Piezoelectric and Piezoresistive} devices measure the
force by exploiting, respectively, the voltage or the
change in resistance to which a body of a specific material is subject when deformed.
Piezoelectric transducers can be used to build both emitters and receivers,
while the use of piezoresistive materials is restricted to receivers.\\

\textbf{Capacitive} devices exploit a condenser whose
capacity changes based on the exerted force. Such a capacity can be turned
into a voltage by charging the capacitor with a known charge.\\

\textbf{Electromagnetic} devices use a coil of conductive material
wrapped around a moving cylindrical magnet that is left free to slide
along the coil's axis. Applying a current to the coil induces a magnetic field
that in turn moves the magnet. Electromagnetic force transducers can be used both
as emitters and as receivers.

\subsection{Light}\label{sec:basic_light}

Light receivers or emitters are used to construct sensors that measure the
amount of light radiation in a certain region of the environment or
to directly determine the geometry of the environment by means of distance
measures to closest objects.  The first task is usually accomplished by assembling 
arrays of light receivers in a one- or two-dimensional matrix to constitute 
the sensitive element of a camera.  The second task is typically done
by either measuring the round-trip time of a beam of light or by processing 
the return of a known light pattern radiated in the environment by an emitter
and sensed by a receiver.\\

\textbf{Photoresistors} are light receivers whose electrical
resistance changes depending on the amount of light radiation to
which they are exposed. They can be designed so that they are
sensible only to a certain spectrum of the optical wavelengths. An
electrical signal is obtained by measuring this resistance of the
device. This is typically done by measuring the voltage drop at the
end of the photoresistor when mounted in a voltage divider
configuration. Typical photoresistors exhibit a latency of around 10
ms.\\

\textbf{Photodiodes and CMOS imaging sensors} are receivers consisting of semiconductor
devices that can generate a current dependent on the amount of light
radiation hitting them. When used in sensors they are typically
driven in reverse configuration (i.e., the cathode is driven positive with respect
to the anode), where they exhibit a linear relationship between
the current and the illuminance within a certain wavelength
spectrum. Integrated arrays of photodiodes and amplifiers couples 
(photosites, or \textit{pixels}) constitute the typical Complementary
Metal-Oxide Semiconductor (CMOS) imaging sensors in consumer cameras.\\

\begin{figure}[t!]
   \centering
   \includegraphics[width=\columnwidth]{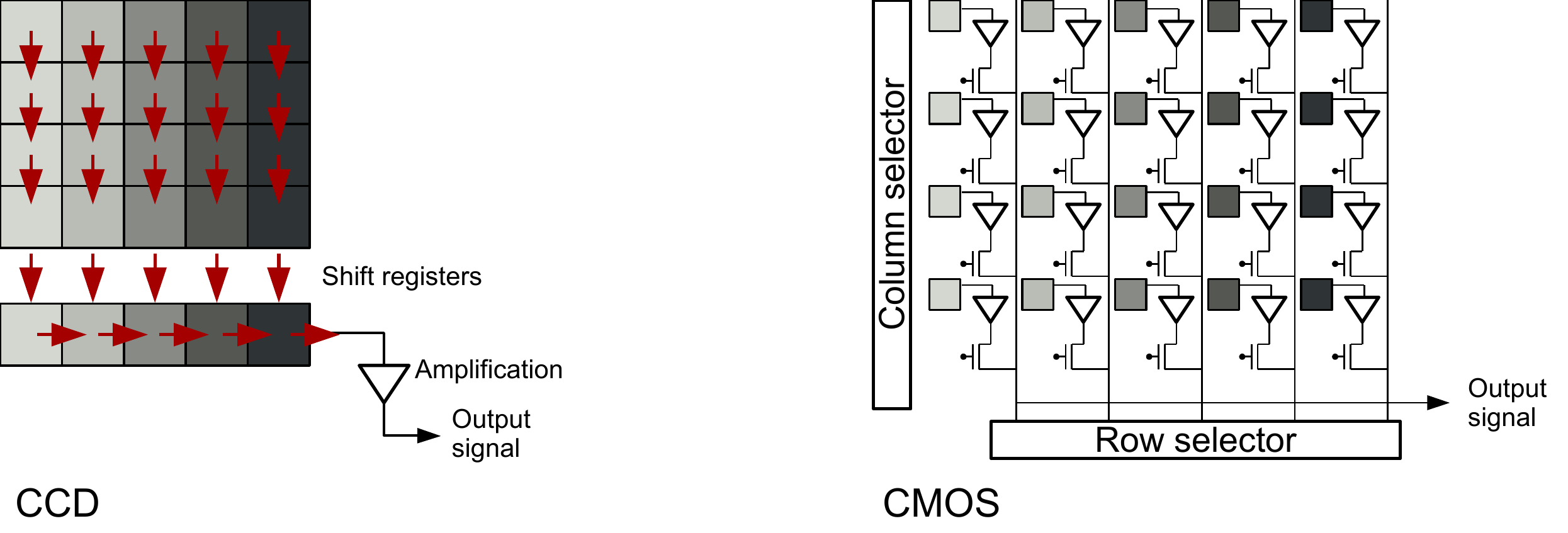}
   \caption{Overview of the popular CCD and CMOS imaging sensors, framing the same image 
   (a gray level linear gradient). In a CCD, the acquired signals are shifted between 
   photosites within the device, amplification is performed one pixel at a time. In CMOS
   sensors, each photosite includes its own amplifier. In both cases, the output signal is 
   usually converted to digital by an ADC.}
   \label{fig:ccd_cmos}
\end{figure}

\textbf{Charge Coupled Devices (CCD) imaging sensors}  are semiconductor light
receivers typically used in imaging sensors.  A CCD appears
electrically as an array of light-sensitive capacitors (as in CMOS sensors, 
\textit{pixels}) that accumulate a charge proportional to the amount of light 
hitting them in a time interval (integration period). The amplification and reading process 
is performed sequentially, one pixel at a time, thanks to shift registers. 
Diagrams of the basis structures of both the CCD and CMOS imaging sensors are
reported in \figref{fig:ccd_cmos}.\\

\textbf{Light Emitting Diodes (LED)} are semiconductor light
emitters that produce light radiation when traversed by a current.
An individual LED can emit light only in a specific light spectrum,
that depends on the doping of its silicon layer.\\

\textbf{Lyotropic Liquid Crystals} are materials whose molecular
configuration changes depending on the current traversing them. A
change in the configuration results in a change in the transparency
to specific light wavelengths.  Usually liquid crystals are
assembled in arrays to form light filters used in conjunction with a
light emitter to construct projectors or LCD displays.

\subsection{Electromagnetic Field}\label{sec:basic_electromag}

Electromagnetic transducers are more commonly known as antennas. They
are composed of an array of conductors. When this array is exposed
to an electromagnetic field, it produces a current dependent on the
electromagnetic field to which it is exposed.  Antennas can be built
to have different directionality, and they can be used both as
emitters and as receivers.  A common use of antennas is in GNSS
(Global Navigation Satellite Systems) sensors, where they are used
as receivers to sense the signal emitted by the satellites. 
Radar (see \secref{sec:radar}) represents another use of antennas
where they are used in a highly directional emitter/receiver
configuration to determine the range of an object by measuring the
round-trip-time of an electromagnetic pulse.

\subsection{Magnetic Field}\label{sec:basic_magnetic}
Magnetic field in mobile robotics is exploited mostly to sense the direction
of the magnetic North. Magnetic field receivers are called magnetometers and 
can rely on one of the following technologies.\\

\textbf{Hall Effect} magnetometers consist in a conductor that is
traversed by a current. When subject to a magnetic field, the
conductor exhibits a voltage in the direction orthogonal to the
current traversing it.  The intensity of the magnetic field along
the direction orthogonal to the flow of current can be measured
through this voltage.\\

\textbf{Magnetoresistive} magnetometers are made of stripes of NiFe
magnetic film, whose resistance changes when exposed to a magnetic
field. They have a well-defined axis of sensitivity.\\

\subsection{Converting Signals}

All the receivers mentioned above generate an electric
signal. Depending on the type of sensor, one is interested in
measuring different characteristics of the signal. The most prominent
are the magnitude, the time at which an impulse arrives, or the phase
of a periodic signal.\\

The \textbf{Magnitude} is converted into a digital number through an 
analog-to-digital converter (ADC). This process is subject to an 
unavoidable quantization error that occurs when a continuous quantity 
is discretized.  The time required for conversion and the noise of the 
resulting measure greatly depends on the type of ADC used.\\

The \textbf{Time} at which an impulse is received is usually captured by
a comparator coupled with a free-running, high-frequency counter. When
the impulse is received, the counter is stopped and its value is
read.  The resolution of these devices depends on the counter's
frequency. The input signal is typically preprocessed by a filtering
stage that avoids spurious readings due to noise in the input.\\

The \textbf{Phase difference} between two periodic signals is the
fraction of a cycle that separates the two in phase from the complete overlap.
The phase is indirectly used to determine the time
difference between the two signals. 


\section{Proximity and Contact Sensors}\label{sec:proximity}

A proximity sensor is able to sense objects placed in 
front of it or nearby, at a fixed or parameterizable distance.
If the distance is zero, it is called a contact sensor. 
These sensors are often used in mobile robots for safety functions, 
for example, to implement emergency stop behaviors in 
case of unexpected obstacles; they are also used as navigation sensors 
in simple mobile robots as robot vacuum cleaners.\\

\begin{figure}[h]
\centering
\begin{minipage}[b]{0.3\columnwidth}
\includegraphics[width=\linewidth]{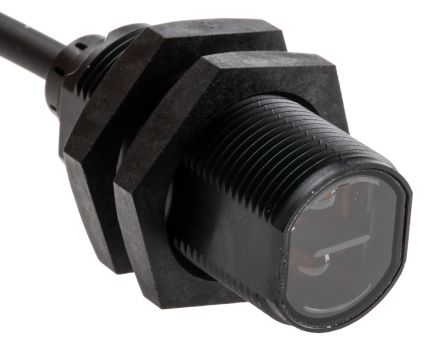}
\end{minipage}\hfill
\begin{minipage}[b]{0.4\columnwidth}
\includegraphics[width=\linewidth]{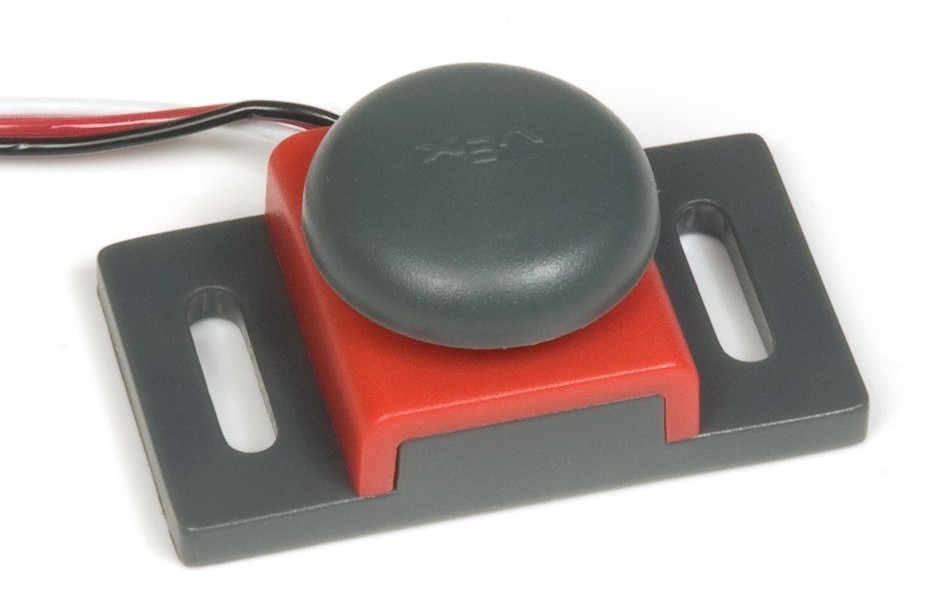}
\end{minipage}\hfill
\caption{Examples of an Omron E3FA-D diffuse-reflective proximity sensor (left) and 
a VEX Robotics bumper (right).}   
\label{fig:proximity_bumper}
\end{figure}

\textbf{Photoelectric proximity sensors} (e.g., \figref{fig:proximity_bumper}, left), also called
diffuse-reflective sensors, are typically composed of an emitter/receiver couple embedded 
in the same case. An emitter (e.g., a \textit{LED}, \secref{sec:basic_light}) transmits a light radiation with a defined 
wavelength, direction, and beam angle; a receiver placed in the line-of-sight
of the emitter (e.g., a \textit{photoresistor} sensitive to the specific wavelength, \secref{sec:basic_light}) 
senses the light eventually reflected back from a nearby object. 
These devices usually output a voltage corresponding to the detected distance, or a digital output that changes if the distance drops below a fixed distance threshold.\\

\textbf{Bumpers} (\figref{fig:proximity_bumper}, right) are simple contact sensors 
that detect a physical collision by means of the pressure or release of a 
microswitch attached to a protective case designed to receive shocks.\\

\section{Encoders}\label{sec:encoders}

An encoder is a proprioceptive device that converts a linear position (\textit{linear encoders})
or an angular position (\textit{rotary encoders}) into a digital code. The latter type is widely
used in mobile robotics, to measure the angular position of the robot wheels. From this information, 
knowing the wheels' diameter and track, it is possible to derive information such as the linear 
and rotational velocities of the robot, and consequently an estimate of its relative motions 
in a planar environment. The most important type of rotary encoders are:\\

\begin{figure}[ht!]
   \centering
   \includegraphics[width=\columnwidth]{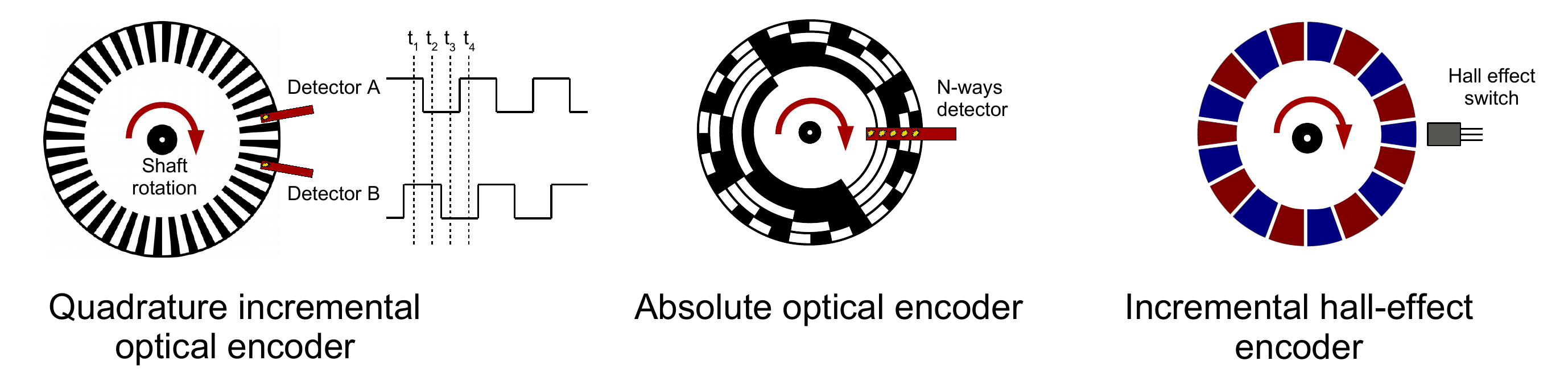}
   \caption{Overview of three types of encoders.}
   \label{fig:encoders}
\end{figure}

\textbf{Optical encoders}: in their simpler implementation, they are composed of two 
opposite transducers: a light emitter  (e.g., a LED) and a photoreceptor (see \secref{sec:basic_light}). 
Between the transducers pair, it is placed a disk (e.g., \figref{fig:encoders}, left),
fixed to the shaft, with a sequence of transparent  (white in  \figref{fig:encoders})
and opaque areas. Depending on the angular rotation of the shaft, the photoreceptor
can detect or not the light diffused  by the LED, providing in output a square wave.
Incremental encoders just provide  information about the motion of the shaft, by counting
the periods (\textit{tics}) of such wave. The quadrature incremental encoder improves this
design  by employing a second emitter-detector pair, shifted in phase of 90 degrees with
respect to the first square wave. Combining the two output waves, it is possible to 
increase the resolution by 4 times while detecting the direction of rotation.
Absolute encoders (e.g., \figref{fig:encoders}, center) also provide the absolute 
position of the shaft by employing multiple  concentric discs and detectors: 
for each tic, the disc provides a unique binary code.\\

\textbf{Hall-effect encoders}: they are composed of a sequence of north-south magnetic poles,
equally spaced and arranged in a circle (e.g., \figref{fig:encoders}, right). A Hall effect 
magnetometer (see \secref{sec:basic_magnetic}) placed near the disc is used to measure the magnetic
field. Often, the  magnetometer is combined with a threshold detector, to provide in output
a square wave: in this case, the operating principle is very similar to the one presented for the
optical encoders.

\section{Global Positioning Systems}

\begin{figure}[ht]
   \centering
   \includegraphics[width=\columnwidth]{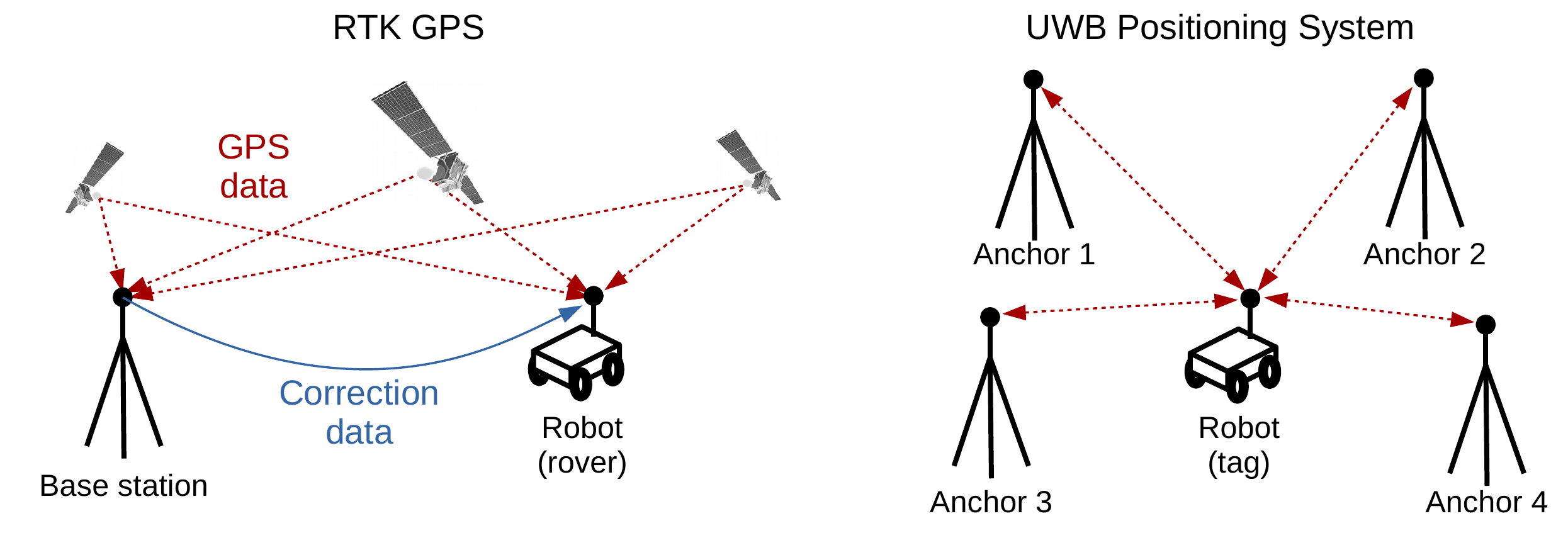}
   \caption{Operating principle of RTK GPS system (left); a simple ultra-wideband positioning system (right).}
   \label{fig:gps_uwb}
\end{figure}

A simple way to localize a mobile robot is to use a set of localized 
and distinguishable landmarks or beacons, spread in the environment 
where the robot moves. By measuring the distance and/or the bearing from one or 
more beacons, the robot can infer its global position. Global Navigation Satellite 
Systems (GNSSs) and Ultra-wideband (UWB) positioning systems build upon this
principle to provide the robot with positioning in outdoor and indoor environments.

\subsection{Outdoor Positioning: Global Navigation Satellite System Receivers}\label{sec:gps}

In GNSSs, the set of beacons is represented by a satellite constellation orbiting
in different orbits, to ensure global coverage. Each satellite constantly and
synchronously transmits its identity, transmission time, and position. 
A GNSS receiver can sense signals (see also \secref{sec:basic_electromag})
from a subset of satellites and, by multiplying the speed of light by the time elapsed 
from the transmission time, it can estimate the distance to each sensed satellite 
(often called ''pseudorange``) hence, by using trilateration techniques, 
its position, represented by its longitude, latitude, and 
elevation. GNSS receivers can't work indoor or in the presence of 
occlusions that do not allow to receive enough satellite signals. 
Among the GNSSs, the American Global Positioning System (GPS) 
is by far the most used. Improvements of this system include:\\

\textbf{Differential Global Positioning Systems} (\textit{DGPS}), where the GPS receiver communicates
with one or more geo-located reference (base) stations to correct the common errors in the measured pseudoranges 
(e.g., due to clock drifts, transmission delays in the ionosphere, etc.).\\ 

\textbf{Real-time kinematic} (\textit{RTK}) positioning technique (see \figref{fig:gps_uwb}, left),
which uses as input both the information sent by the satellites and the carrier waves of the
received signals. The distance with respect to a satellite is basically calculated by multiplying 
the wavelength of the carrier by the number of full cycles between the satellite and the receiver
and adding the phase difference. As in DGPS, RTK systems exploit a base station that 
transmits the phase of the carrier it observes to all rovers (e.g., mobile robots)
to provide real-time corrections. Correction data is typically transmitted 
by using a radio modem, for example in the UHF band. RTK receivers, coupled with a 
geo-located reference station, enable to improve the localization accuracy from a nominal GPS accuracy which, depending on the environment, ranges from 5 to 15 meters, to centimeter-level accuracy.

\subsection{Indoor Positioning: Ultra-wideband Positioning Systems}\label{sec:uwb}

Ultra-wideband refers to a technology for transmitting radio signals through
the use of extremely short energy pulses (duration nanoseconds to microseconds) 
and therefore with very wide spectral occupation (frequencies between 3.1 and 10.6 GHz).
UWB can provide a high data transfer rate with low power consumption.
This technology is used in real-time positioning systems, relying on concepts similar to GNSS.
Typically a set of UWB beacons called \emph{anchors} is distributed in the environment at fixed,
known locations (see \figref{fig:gps_uwb}, right), while a UWB \emph{tag} is installed
on the agent to be localized (e.g., a mobile robot). Unlike GNSS, communications 
are bidirectional. Basically, UWB positioning systems rely on two alternative
algorithms:\\

\textbf{Time Difference of Arrival} (TDOA): The UWB tag should constantly broadcast a message
stating its identity; nearby anchors will receive it at different times as their positions
(and distances) are different. If the clocks of the anchors are synchronized, it is possible
to use these time differences to retrieve the position of the tag through triangulation.\\

\textbf{Time of Flight} (TOF): To estimate the ranges between UWB devices in the absence of a global clock, a two-way protocol is used to measure the round trip time of messages. 
A message is sent by the UWB tag to the anchors and immediately bounced back from them: from the flight time of the signal, the distance between the anchor and the tag can then be measured, hence the tag position can be estimated through triangulation.\\

The orientation of the tag can be achieved by using a three-dimensional arrangement of its antennas. When the wavefront of the signal passes through the antennas system, it will hit each antenna at a different time, depending on its direction.
This direction can be obtained from the arrival time differences, by resolving a linear system of equations. Since the time interval is typically extracted from the phase difference between a pair of antennas and this interval can be larger than the wavelength, wraparounds require to be taken into account.\\

UWB positioning systems can typically achieve sub-meter accuracy and, for some special use cases, a few centimeters accuracy. They can be used both in outdoor and indoor environments but, since they require adequate anchors coverage, the latter are the typical context of application.
Although attractive, the use of UWB presents some issues since signals at this frequency are absorbed by bodies with high water content and deflected by metal structures; hence to be reliable, several anchors need to be used.

\section{Inertial and Heading Sensors}

Inertial sensors are based on the principle of inertia, i.e. the resistance of a body 
to change its current state of stillness or motion with constant linear or 
rotational velocity. The most common inertial sensors are the accelerometers
and gyroscopes. They are often combined together inside sensor compounds 
called  Inertial Measurement Units (IMUs), often along with a heading sensor 
such as a compass. 

\subsection{Accelerometers}

\begin{figure}[h]
\centering
\begin{minipage}[b]{0.4\columnwidth}
\includegraphics[width=\linewidth]{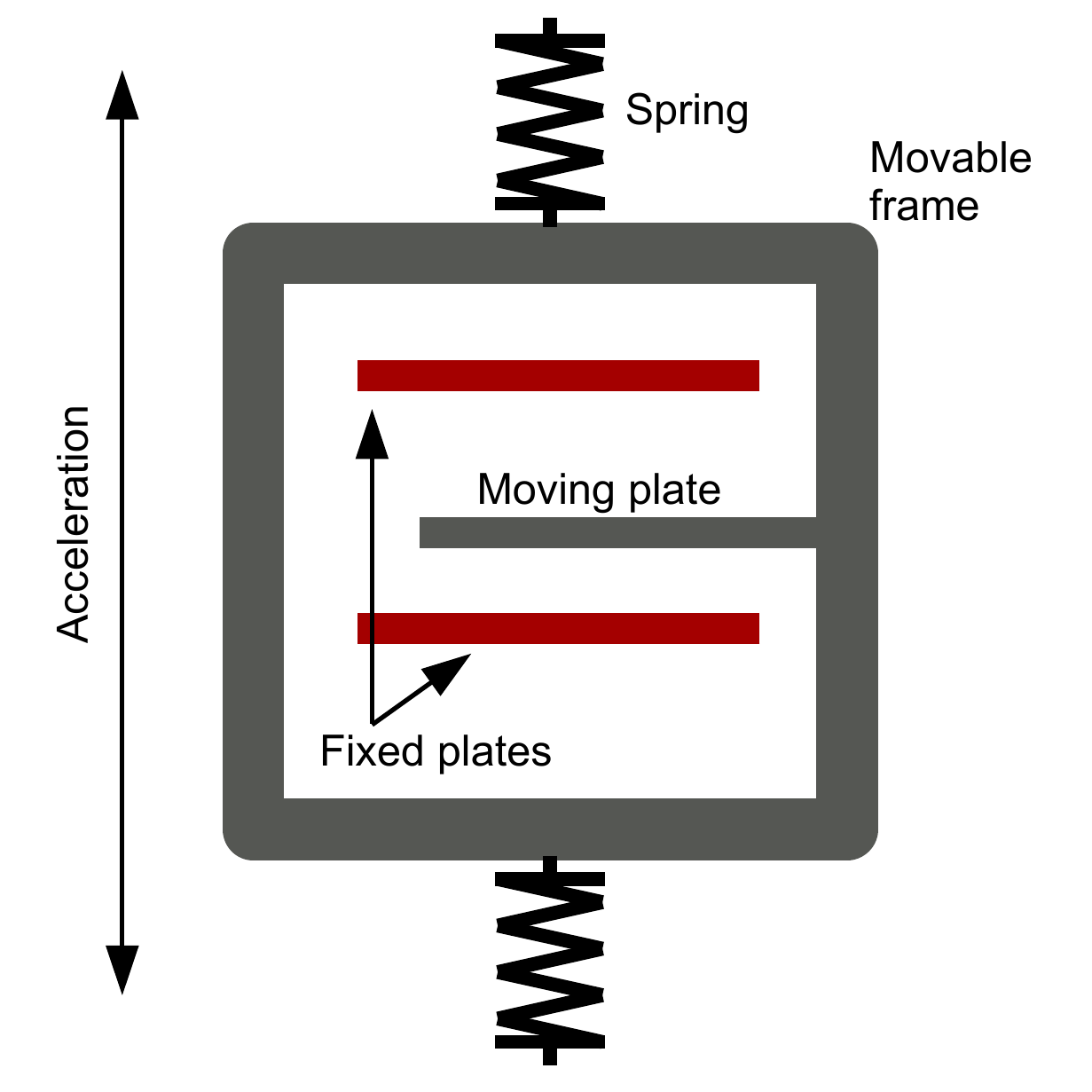}
\end{minipage}\hfill
\begin{minipage}[b]{0.5\columnwidth}
\includegraphics[width=\linewidth]{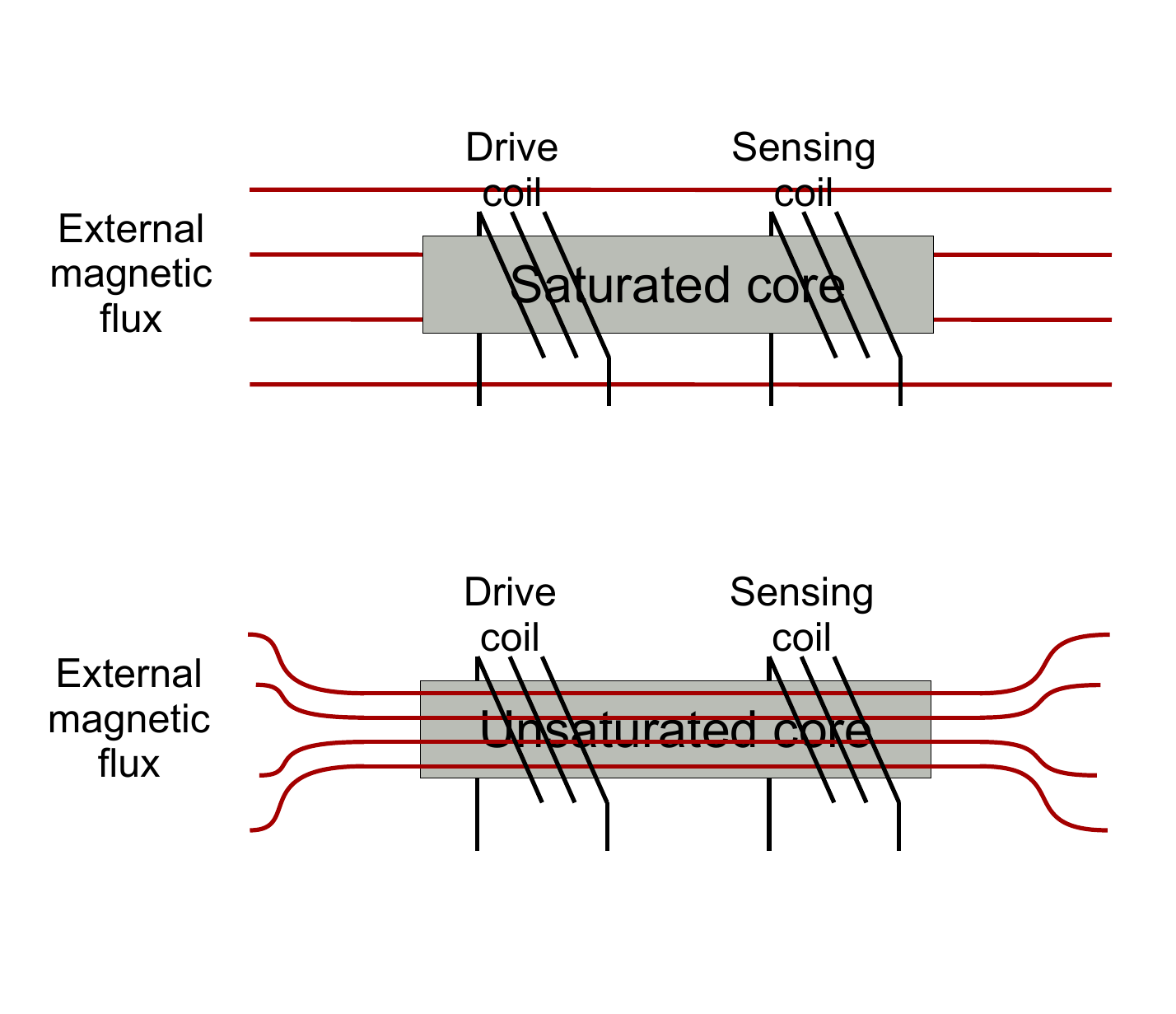}
\end{minipage}\hfill
\caption{A basic MEMS accelerometer sensing unit (left);
         unsaturated and saturated magnetically permeable core in fluxgate compasses (right).}   
\label{fig:acc_compass}
\end{figure}

Accelerometers are devices that measure the linear acceleration 
they undergo along a defined axis. The acceleration is typically converted into 
either a force or a deformation (see \secref{sec:basic_force}) through some mechanical
construction. When operating on the Earth, the gravitational acceleration is usually the dominant
component in the acceleration vector of a body, and its direction,
as well as magnitude, plays a crucial role in having reliable measurements.\\

\textbf{Piezoelectric} accelerometers, in their basic implementation, build upon 
a mass-spring-piezoelectric material stack: the mass-spring system converts the acceleration
in a force (i.e., a displacement), such force is then measured by the 
piezoelectric element by means of change in voltage or in resistance.\\

\textbf{Capacitive} accelerometers use a system like the one shown in 
\figref{fig:acc_compass} (left), composed of a moving element with not negligible mass 
tethered by tiny springs to the accelerometer frame. Both the moving and fixed elements
are equipped with plates to form a variable capacitor circuit. When experiencing
an acceleration, the moving frame changes its position with respect to the fixed
accelerometer frame, so resulting in a change of capacity that can be turned
into an electrical signal that is proportional to the sensed acceleration. 
MEMS accelerometers typically use such configuration as basic sensing unit.

\subsection{Gyroscopes}

Devices used to measure the orientation are commonly known as gyroscopes. 
The most common principle is to exploit the inertia of a
rotating or vibrating body, which tends to preserve the initial body's orientation.
More expensive fiber-optic gyroscopes rely on the Sagnac effect on light and are
in fact ensembles of different emitters, receivers, and processing
devices.\\

\textbf{Rotating Structure Gyroscopes} use a spinning disk mounted on
a structure that allows the disk to preserve its orientation when
the support is rotated. Thanks to the rotational inertia, the disk
preserves its absolute orientation.  Measuring the orientation of
the disk with respect to the structure provides the relative
rotation from when the gyroscope was first started.  Frictions in
the structure might result in a loss of accuracy over time. These
types of transducers are not used in common mobile robots, but
presented here for completeness.\\

\textbf{Vibrating Structure Gyroscopes (VSG)} exploit the fact that
a vibrating object tends to vibrate along the same plane, even if
the orientation of its support changes. When the support is rotated,
the vibrating structure exerts a force on the support.  Measuring
this force the \emph{rate} of rotation can be
measured. Alternatively, if the plane is left free to move one can
measure its inclination. Multiple sensors mounted with non-co-planar
orientations allow to estimate the full three-dimensional rotational
velocity vector.  Piezoelectric and MEMS gyroscopes fall into the
VSG category. These types of gyroscopes do not directly provide
absolute orientation, but the rotational rate. Absolute
orientation can be obtained by integrating the rotational
rate over time. \\

\textbf{Fiber-Optic Gyroscopes (FOG)} exploit the Sagnac effect that
results in the interference between two beams of light that depends
on the rotational rate. The interference figure generated by the
light beams is then related to the rotational rate. Usually, FOGs
convey a laser beam in a winding made of optical fiber
to increase the length traveled by the light and thus magnifying the
interference. Fiber-optic gyroscopes are highly accurate and they can
provide accurate estimates of orientation even after an integration
procedure. This comes at the cost of being substantially more
expensive and bulky than their MEMS counterpart.

\subsection{Compasses}

Compasses are sensors for measuring the direction of the Earth's magnetic  
field.\\

\textbf{Hall Effect} compasses used in mobile robots are typically implemented using 
two or three perpendicular Hall Effect magnetometers (see \secref{sec:basic_magnetic}).\\

\textbf{Fluxgate compasses} are more expensive devices that however can provide superior accuracy.
The basic element of a fluxgate compass is a magnetically permeable core wound by two
coils of wire (\figref{fig:acc_compass}, right): one is used to induce a magnetic field (drive coil), 
the other to measure the magnetic field (sensing coil). If the induced magnetic field saturates the 
permeable core, the flux of any external magnetic field (e.g., the Earth's magnetic field) will be 
unaffected by the presence of the saturated core, and the sensing coil will measure only the auto-induced 
field. Conversely, the external flux tends to pass toward the permeable core, inducing an electromotive 
force that can be measured by the sensing coil. By integrating two perpendicular elements of this type,
and alternating their saturation and unsaturation states, it is possible to compute the direction 
of the external magnetic field by taking into account the phase differences with respect to the induced 
magnetic field.
\subsection{Inertial Measurement Units (IMUs)}\label{sec:imus}

\begin{figure}[h]
\centering
\begin{minipage}[b]{0.4\columnwidth}
\includegraphics[width=\linewidth]{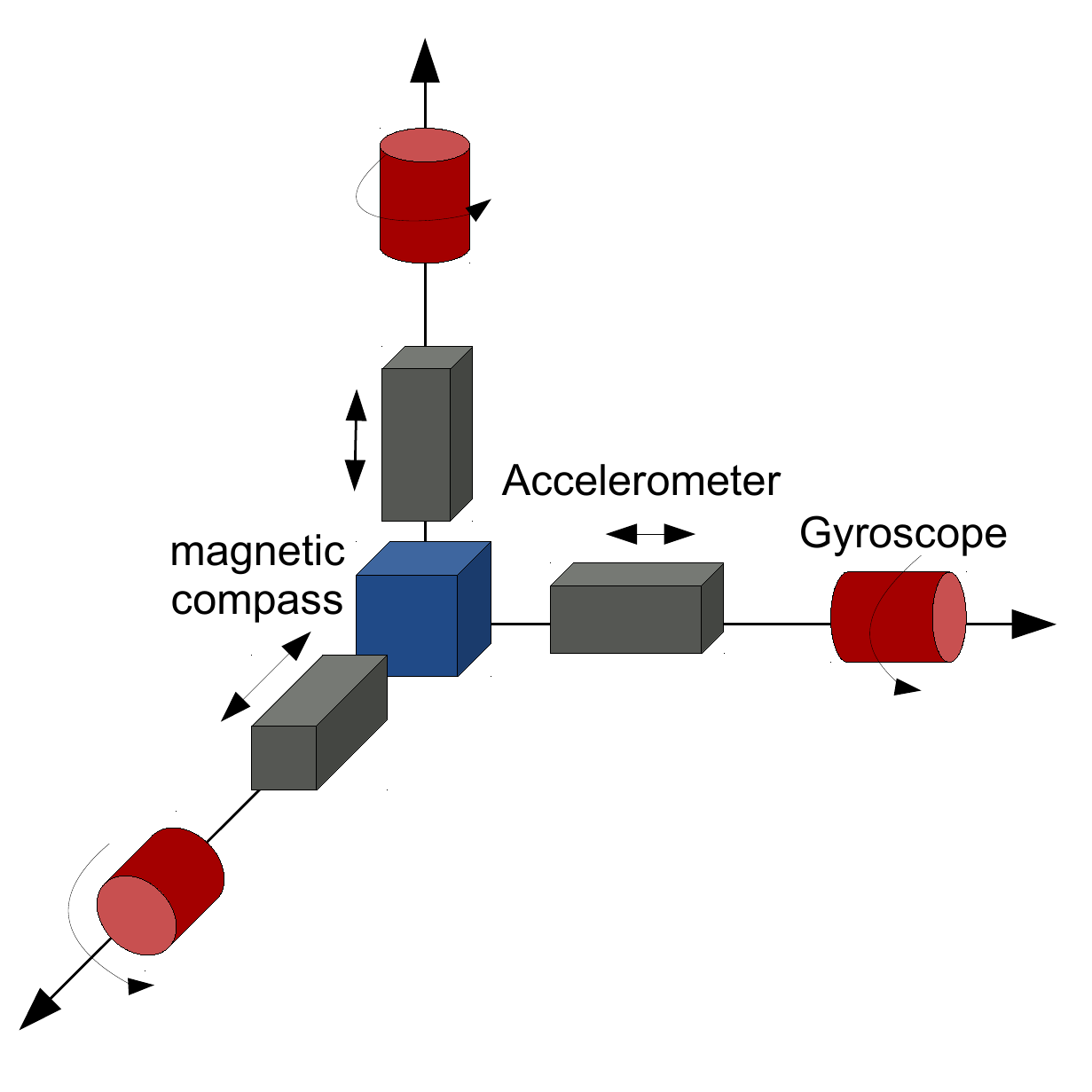}
\end{minipage}\hfill
\begin{minipage}[b]{0.5\columnwidth}
\includegraphics[width=\linewidth]{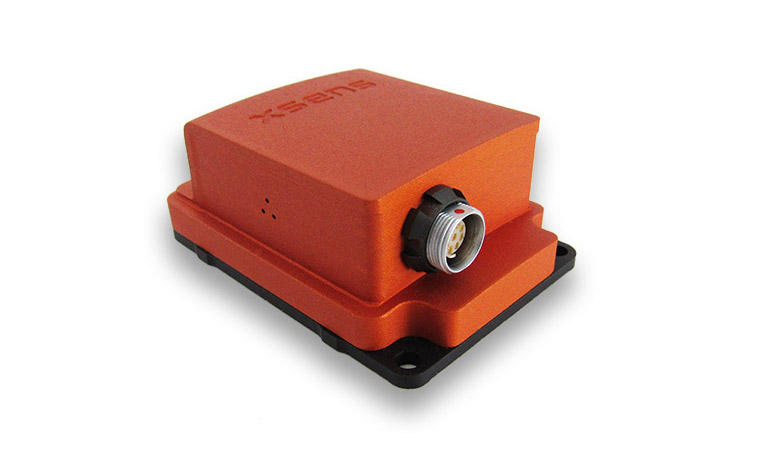}
\end{minipage}\hfill
\caption{Conceptual arrangement of sensors within an IMU (left); the XSens MTi-300 AHRS IMU (right).}   
\label{fig:imu}
\end{figure}

IMUs are composed of a tri-axial cluster of accelerometers and 
a tri-axial cluster of gyroscopes, usually based on MEMS technology. 
The two triads define a single,  shared, orthogonal 3D frame, and they are often associated with a 
magnetometer (\figref{fig:imu}, left). 
An AHRS (Attitude and Heading Reference System, e.g., \figref{fig:imu}, right) IMU provides
3D orientation by internally integrating the gyroscopes and  fusing this data with the accelerometer 
and the magnetometer data. IMUs are often used in mobile robotics to integrate and 
to improve the consistency and the reactivity of the robot navigations systems,
e.g., in GNSS-based navigation systems.

\section{Digital Cameras}\label{sec:cameras}

\begin{figure}[h]
\centering
\begin{minipage}[b]{0.4\columnwidth}
\includegraphics[width=\linewidth]{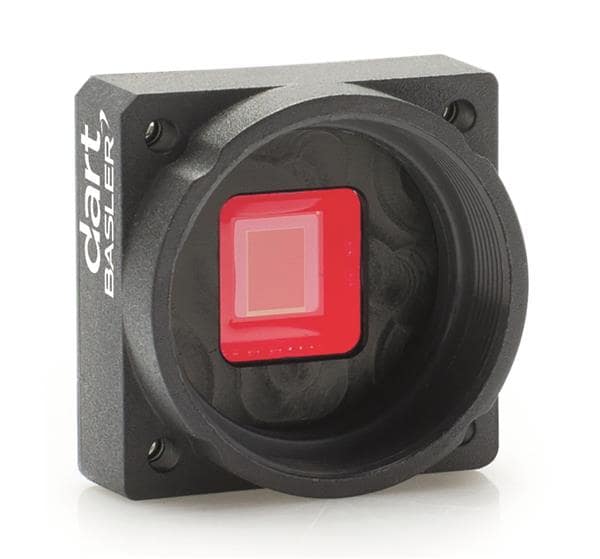}
\end{minipage}\hfill
\begin{minipage}[b]{0.5\columnwidth}
\includegraphics[width=\linewidth]{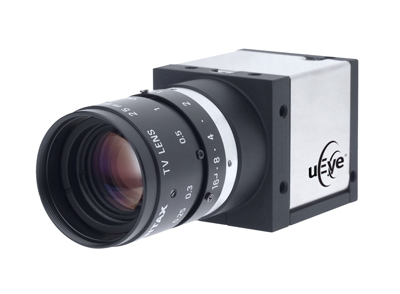}
\end{minipage}\hfill
\caption{Two examples of digital, area scan cameras. A Basler Dart without lens, 
with visible CMOS imaging sensor (left): an IDS uEye with mounted lens.}   
\label{fig:digital_cameras}
\end{figure}

\begin{figure}[h]
   \centering
   \includegraphics[width=\columnwidth]{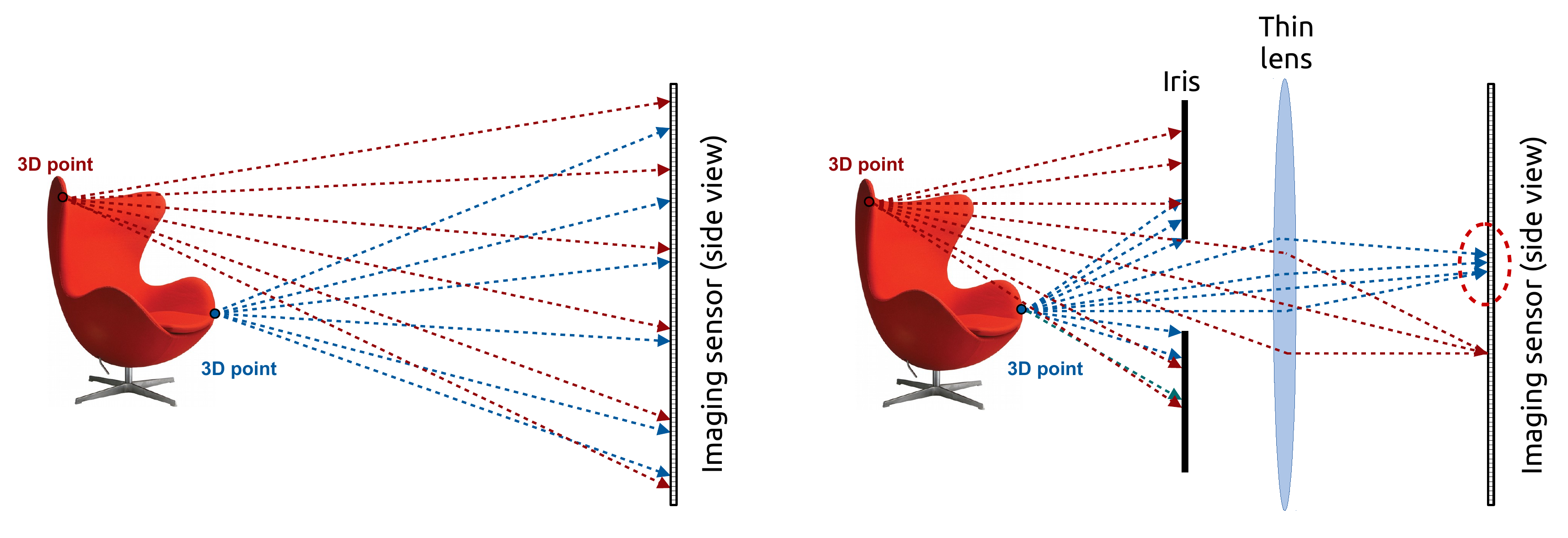}
   \caption{Without a lens, each 3D point of the scene will project 
   its radiation into each point of the 2D imaging sensor, producing 
   a useless image (left); A lens in front of the sensor (right), here 
   represented by its simplest model, the \textit{Thin lens model}, is used to 
   mitigate as possible this phenomenon.}
   \label{fig:thin_lens}
\end{figure}

Digital cameras (also called \textit{area scan cameras}) are devices 
able to produce two-dimensional (2D) arrays (called \textit{images}) 
of measurements of a specific electromagnetic radiation (e.g., the visible light) 
coming from non-occluded surfaces of a framed three-dimensional (3D) scene. 
In this 3D-2D projection, one dimension (i.e., the points' depth) 
is anyhow lost.\\
The basic components of a digital camera are (a) an imaging sensor, 
e.g., a CCD  or a CMOS (see \secref{sec:basic_light}, e.g., \figref{fig:digital_cameras}, left); 
(b) an optical system, called \textit{lens}, used to
route the sensed information (e.g., the  visible light) from each 3D point
toward a 2D point of the imaging sensor (e.g., \figref{fig:digital_cameras}, right); 
(c) an internal ADC used to convert each 
pixel measurement into a digital value.\\
In nature, each 3D point constantly emits radiations that are spread in 
all directions in the space, possibly reaching the whole area of an 
imaging sensor exposed to such radiations without a lens 
(see \figref{fig:thin_lens}, left). In the ideal case, for each 3D point emitting 
such omnidirectional radiations, there should be exactly one 2D point of the 
imaging sensor that receives a single ray of this and only this radiation. 
This ideal model is called \textit{Pinhole model}. The aim of the lens is to 
approximate the ideal case by routing as possible multiple radiation rays coming 
from the same 3D point  toward a single 2D point of the imaging sensor
(see \figref{fig:thin_lens}, right). This is also obtained by introducing 
a barrier with a central opening (called \textit{Iris}) to block most of the rays.
Although this model (called \textit{Thin lens model}) only approximates
the ideal one (e.g., the lower 3D point of the right side of \figref{fig:thin_lens}
is projected in a neighborhood of points into the 2D imaging sensor), 
this model and especially the Pinhole model are commonly 
used to describe and model real cameras. Cameras
that fall into such model are also called \textit{perspective cameras}.\\
Digital cameras can be used in a very wide range of applications in mobile robotics, 
among others: place recognition, visual servoing, ego-motion estimation, SLAM (Simultaneous Localization and Mapping),
object detection and classification, etc.\\


\begin{figure}[h]
   \centering
   \includegraphics[width=\columnwidth]{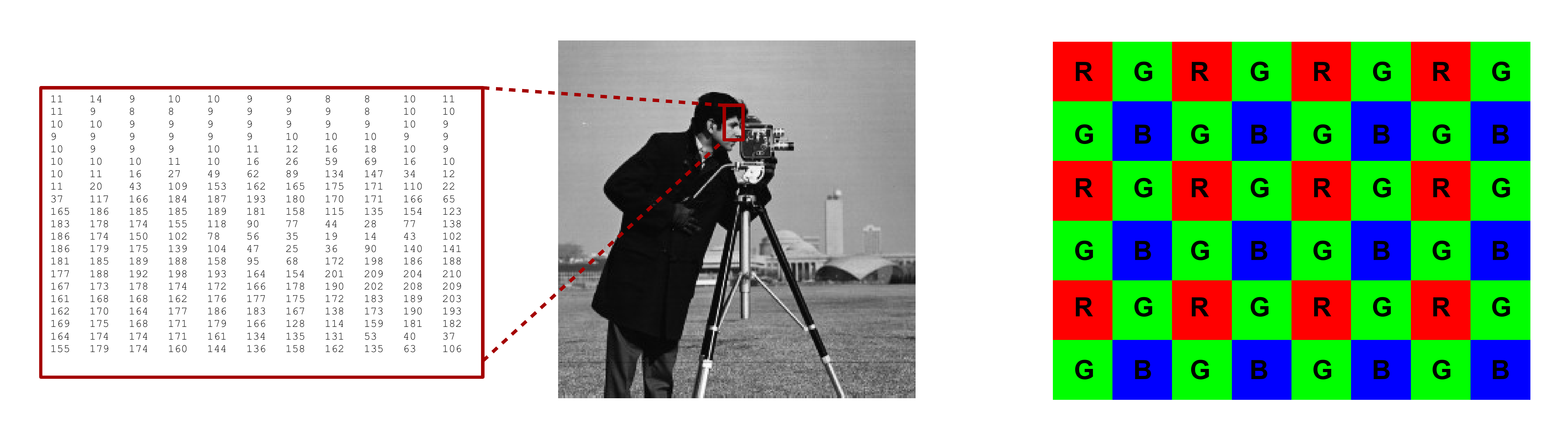}
   \caption{A portion of a single channel, gray level \textit{image} 
   and the corresponding sub-picture (left); Bayer arrangement 
   of an RGB imaging sensor (right).}
   \label{fig:image_picture_bayer}
\end{figure}

\textbf{Gray Level (GL) cameras} are used to simulate the human visual system
by measuring the \textit{intensity} of the electromagnetic radiations 
that lie within the \textit{visible light spectrum}, 
that corresponds to a range of wavelengths from about 380 to about 740 nanometers.
CMOS or CCD imaging sensors used in such cameras are designed to be sensible
only to this spectrum. The provided output is a \textit{brightness} image:
low radiance 3D points will be mapped into small pixel values, saturating
toward the zero value; conversely, high radiance points will be mapped into 
high pixel values, saturating toward the largest integer representation of the
digital sensor. To be easily interpreted by the human eye, an image can be 
represented by a \textit{picture}, that maps each pixel value into a visible 
''color``, in this case a gray level (e.g., \figref{fig:image_picture_bayer}, left).\\

\textbf{Color cameras} extend the functionality of GL cameras by providing 
\textit{color vision} capabilities, i.e. by measuring also the \textit{wavelength}
of the electromagnetic radiations spread by 3D points. Color cameras are often called 
\textit{RGB cameras} since they usually employ the \textit{RGB additive color model},
that represents the color information by means of three-channel images. 
Color cameras imaging sensors include three types of photoreceptors; each type is sensible 
to a specific wavelength spectrum centered around one of the three primary colors 
of the RGB model: red, green, and blue. For each pixel, such cameras provide 
a triad of values corresponding to the intensity of these three colors,
so reproducing a wide range of colors. An RGB camera can be assembled by 
employing three separate imaging sensors, each one sensitive to a specific
wavelength spectrum, and by a prism that splits each light beam coming
into the lens into three beams that are projected in exactly the same 2D 
locations of each of the three sensors. 
A simpler and more popular implementation employs a single imaging sensor 
in which each pixel is sensitive to a specific wavelength spectrum.
The distribution of the pixels follows a fixed pattern, e.g. the Bayer arrangement
depicted in \figref{fig:image_picture_bayer} (right). In this case, 
each pixel provides a single value, corresponding to a specific spectrum. 
A special interpolation algorithm is then used to recover the R, G, and B 
values for each pixel.\\

Digital cameras can be used also to detect radiations that are not visible to humans, 
such as near-infrared (NIR) radiations. They are basically GL cameras 
sensitive to wavelengths that spread from 750 to 1400 nanometers. 

\subsection{Omnidirectional Cameras}

\begin{figure}[h]
\centering
\begin{minipage}[b]{0.28\columnwidth}
\includegraphics[width=\linewidth]{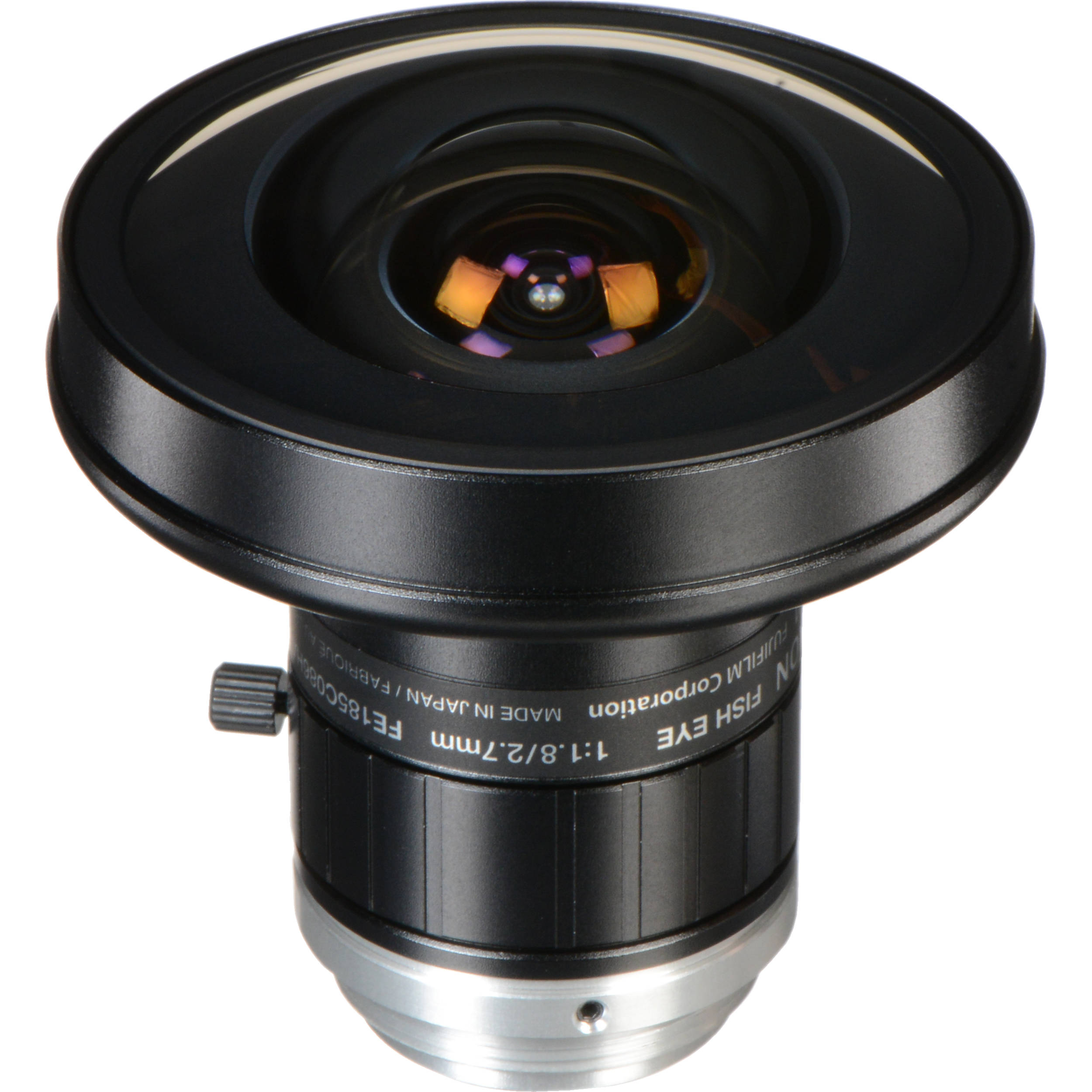}
\end{minipage}\hfill
\begin{minipage}[b]{0.12\columnwidth}
\includegraphics[width=\linewidth]{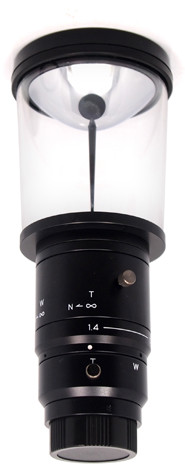}
\end{minipage}\hfill
\begin{minipage}[b]{0.3\columnwidth}
\includegraphics[width=\linewidth]{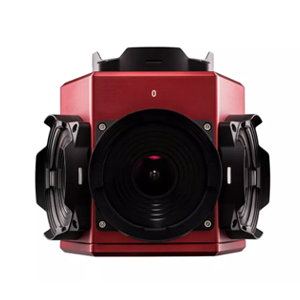}
\end{minipage}\hfill
\caption{A Fujinon fisheye lens (left); a hyperbolic mirror for 
catadioptric cameras (center); the Flir Ladybug 5 polydioptric spherical
camera (right).}   
\label{fig:omnicams}
\end{figure}

Perspective cameras usually are able to frame in one image only a portion 
(\textit{field of view}) of the surrounding scene. This limitation is overcome by 
omnidirectional cameras, able to provide a field of view of at least 180 degrees in 
one or both horizontal and vertical directions. An omnidirectional camera (sometimes called
\textit{panoramic camera}) can be  implemented starting from one or more digital cameras, 
e.g.: (a) by using a \textit{fisheye lens}, that is a lens that provides a very wide 
angle of view  (e.g., \figref{fig:omnicams}, left); (b) by placing in front of the lens a 
special mirror, e.g, parabolic or hyperbolic (\figref{fig:omnicams}, center)
to form a \textit{catadioptric camera}; (c) by assembling an array of cameras 
that frame different points of the scene but have some overlaps between them
(e.g., \figref{fig:omnicams}, right) to form a \textit{polydioptric camera}.\\
Due to the large field of view, an omnidirectional camera can't be modeled by using
the standard Pinhole model: a \textit{spherical projection model}, for instance, 
is a more suitable model to represent such type of cameras.

\subsection{Event Cameras}

Standard cameras sample the scene light synchronously, at a fixed frame rate, e.g., 60 fps (frames per second). In contrast, event cameras are \textit{asynchronous} sensors that measure per-pixel brightness changes and generate a stream of events representing brightness changes. As soon as a pixel detects a brightness change greater than a predefined threshold, an event is sent. Each event is represented by a timestamp, the pixel position, and the sign of change. The output of an event camera depends on the dynamics of the scene: for a static scene, it does not produce any output; for a scene with several moving structures, it produces a number of events depending on the apparent motion.
Similarly to CMOS of standard cameras (\figref{fig:ccd_cmos}), the imaging sensor of event cameras is organized as an array of photoreceptors, but in this case each pixel is coupled with a comparator to detect events and is connected with a shared digital output bus.\\
Even if their use in mobile robots is less immediate compared to traditional cameras, event cameras offer attractive properties such as low latency, high speed, high dynamic range, and reduced motion blur. Event cameras can be used, for example, for obstacle avoidance, object tracking, and SLAM in highly dynamic environments and/or in case of high-speed robot motions.

\section{Ranging Sensors}

Ranging sensors provide the distance to objects on a defined portion
of the surrounding scene. They can be considered an extension of the
proximity sensors (\secref{sec:proximity}) since, compared to the
latter, ranging sensors directly output metric distances to objects,
and provide extended field of view or extended measurement range
(\secref{sec:sensor_specs}). Ranging sensors applied to mobile robots
can be used in several applications, among others robot localization,
SLAM, 3D environment reconstruction, object localization, and obstacle
avoidance.\\ 
Such sensors output arrays of distances
(called \textit{ranges}), each one associated with a defined direction of 
measurement, or equivalently the projection of such distances
(called \textit{depths}) along a fixed direction defined by one of the
axes of the sensor coordinate system. 
A 2D array of ranges is often called \textit {range image}, while a 2D array
of depths is called \textit{depth map}. Cameras that provide a depth map 
are generally called \textit{depth cameras}. From both ranges and depth information, it
is possible to generate a \textit{point cloud} (e.g., \figref{fig:stereo}, right) 
that is a vector of 3D points that represents samples of the 3D structure of 
the surrounding scene. In the following, we briefly present the most common ranging sensors.

\subsection{Sonar and Ultrasonic Sensors}\label{sec:sonar}

\begin{figure}[h]
   \centering
   \includegraphics[width=\columnwidth]{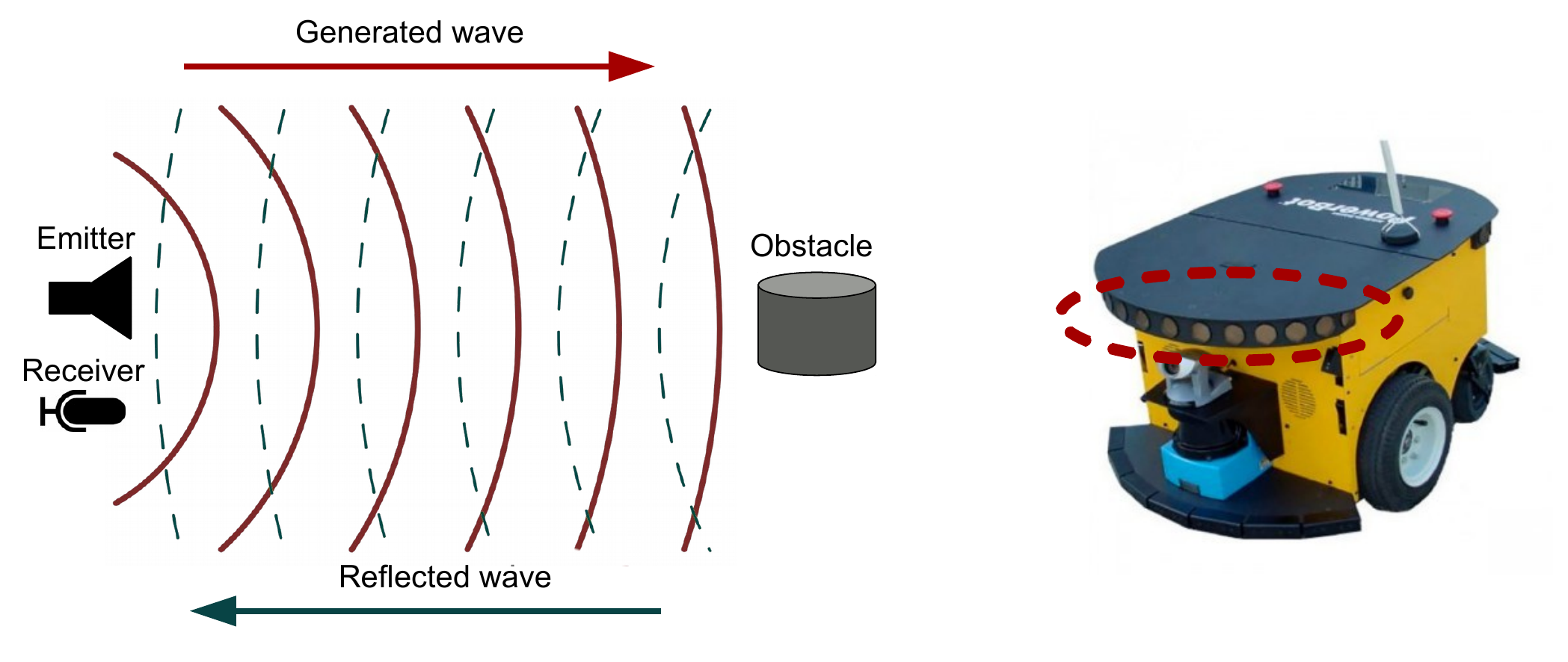}
   \caption{Operating principle of sonars (left); The red dashed ellipse highlight 
   the array of sonars that equip the MobileRobots PowerBot platform (right)}
   \label{fig:sonar}
\end{figure}

Sonar (acronym for Sound Navigation and Ranging) are devices that use sound waves to detect objects and
measure distances. Passive sonar devices only perceive sounds, while active sonars also emit
pulses of sound: the latter are commonly used in mobile robots for obstacle avoidance and
self-localization. Active sonars consist of an emitter/receiver pair
measuring the distance to an obstacle from the round-trip time of
an emitted pulse (\figref{fig:sonar}, left). 
The pressure wave (sound) that is emitted can have different frequencies ranging from very low to high (ultrasound).
The latter are the ones commonly used in sensors for robotics, which are usually called \emph{ultrasonic sensors}. The cone of the emitted signal
can be varied: the narrower the cone, the higher the angular resolution of
the sensor. In order to have multiple readings at different directions,
multiple sonar devices need to be configured in an array (e.g., \figref{fig:sonar}, 
right). Transducers used to build sonars lie in the family of the force and deformation
transducers (\secref{sec:basic_force}).\\

\textbf{Piezolelectric Sound Transducers} rely on materials that
generate a voltage when their shape is changed. This change in shape
occurs as a consequence of sound waves.  They are used both for
detecting sound in the audible and in the ultrasound frequency
bands.  Piezoelectric transducers can also be used to generate a
sound, since piezoelectric materials change shape when subject to a
voltage.\\

\textbf{Capacitive Microphones} are condensers that change their
capacity when exposed to a sound wave.  The charge in the capacitor
results in a voltage that depends on the varying capacity and is
thus related to the pressure exerted by the sound waves on the
transducer's membrane.

\subsection{LiDAR}\label{sec:lidar}

\begin{figure}[h]
   \centering
   \includegraphics[width=\columnwidth]{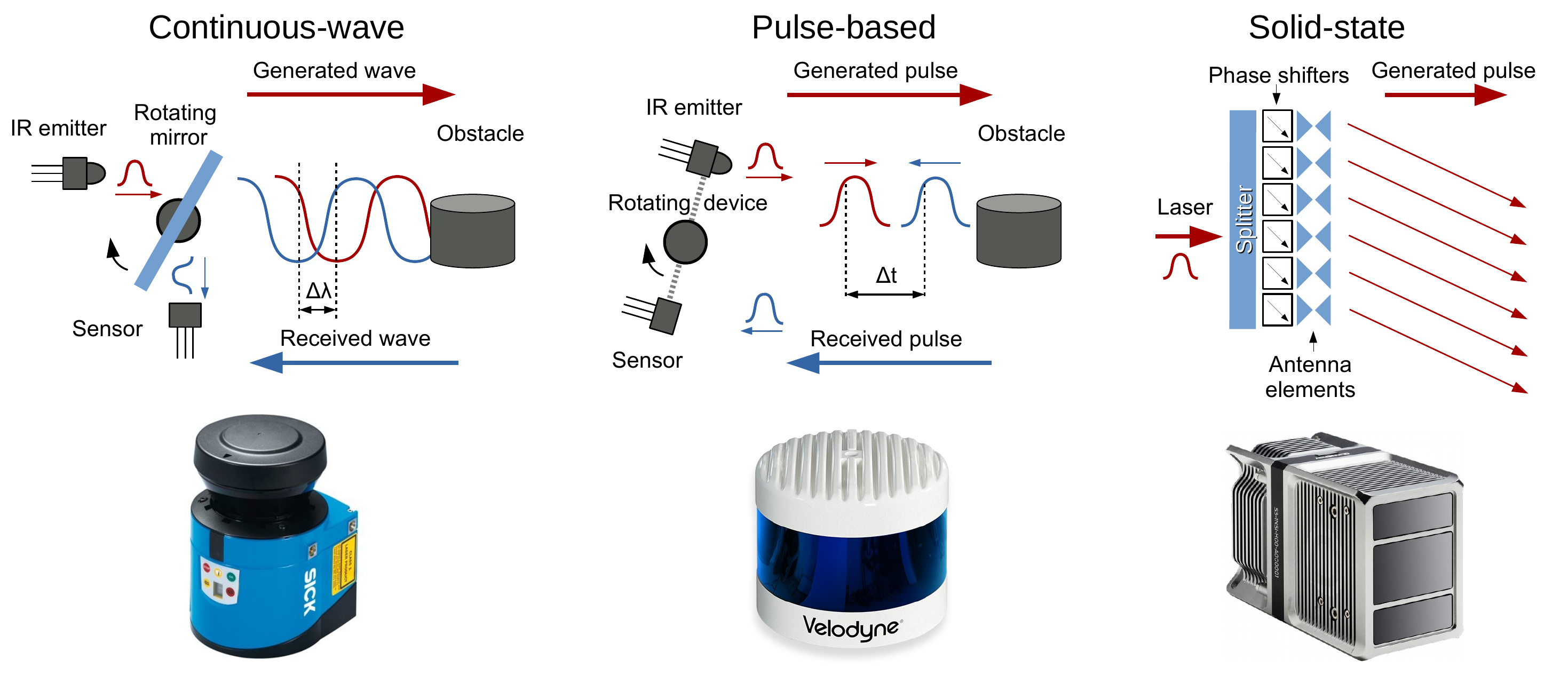}
   \caption{Operating principles (top row) of three types of LiDAR (bottom row): continuous-wave LiDAR with rotating mirror for the SICK LMS 100 (left column); pulse-based LiDAR with spinning laser/detector pairs for the Velodyne Alpha Puck LiDAR that exploits an array of 128 pairs (central column); solid-state LiDAR that uses an optical phased array to deflect laser beams for the Quanergy S3-2 (right column).}
   \label{fig:lidar}
\end{figure}

LiDAR (acronym for Light Detection And Ranging) devices 
exploited in mobile robots use coherent light sources (i.e., \emph{lasers}, 
typically at infrared wavelengths) to measure the distance and reflective
properties of the environment.

There are two main LiDAR technologies, with different working principles:\\

\textbf{Continuous-wave (CW) LiDARs} (also known as phase-shift-based LiDARs) detect
the range by measuring  the \textit{phase difference} $\Delta \lambda$ between a generated 
continuous wave and the reflected wave backscattered by an object encountered along the sensor
line of sight (see \figref{fig:lidar}, left);\\

\textbf{Pulse-based (PB) LiDARs} measure directly the time-of-flight,
i.e. the round-trip-time of a pulsed light (see \figref{fig:lidar}, center). This type needs very short 
laser pulses and high temporal accuracy, considering that the 
speed of light is 0.3 meters/nanosecond.\\

Standard point-based LiDARs can only measure one point at a time per emitting diode.
However, since the time for a single range measurement is
rather short, a sequence of measurements can be rapidly acquired by
deflecting the beam. This is done either by employing a deflecting
mirror (e.g., as in \figref{fig:lidar}, left) or by mounting the transmitting/receiving element on a movable
support that rotates around one or two axes in order to acquire 2D planar scans or 3D range images (e.g., as in \figref{fig:lidar}, center). 
To acquire more data in parallel, \emph{multi-channel LiDARs} use several diodes
pointing in different directions, usually arranged to acquire a planar profile in one shot. 
By rotating or translating the diodes array, it is possible to generate a range image.
Current state-of-the-art systems offer up to 128 pulse-based diodes that can generate 
360-degree range images.\\

\textbf{Solid-state LiDARs} Recently, solid-state LiDARs are entering the market of
sensors for mobile robots.
These sensors allow to obtain a dense 3D scan of the surrounding environment without 
using mechanical tools for deflecting the laser beams. 
Solid-state LiDARs typically use a pulsed laser that feeds an
\emph{Optical Phased Array} (OPA) that is a 2D array of closely spaced
(around 1 µm) optical antennas. 
Variable phase control is applied at each antenna to generate a radiation pattern
that points in the desired direction (see \figref{fig:lidar}, right).\\

Along with sonar arrays, LiDARs are the most commonly used 
sensors in mobile robot navigation and obstacle avoidance.
Compared to sonars, LiDARs have several advantages: (i) increased range;
(ii) higher frequency; (iii) increased accuracy.
The disadvantages of a LiDAR compared to a sonar are higher cost and 
greater sensitivity to fog and dust, although sensors that report multiple 
return echoes are nowadays more common, resulting in greater robustness to adverse events.

\subsection{Radar}\label{sec:radar}

\begin{figure}[ht]
   \centering
   \includegraphics[width=\columnwidth]{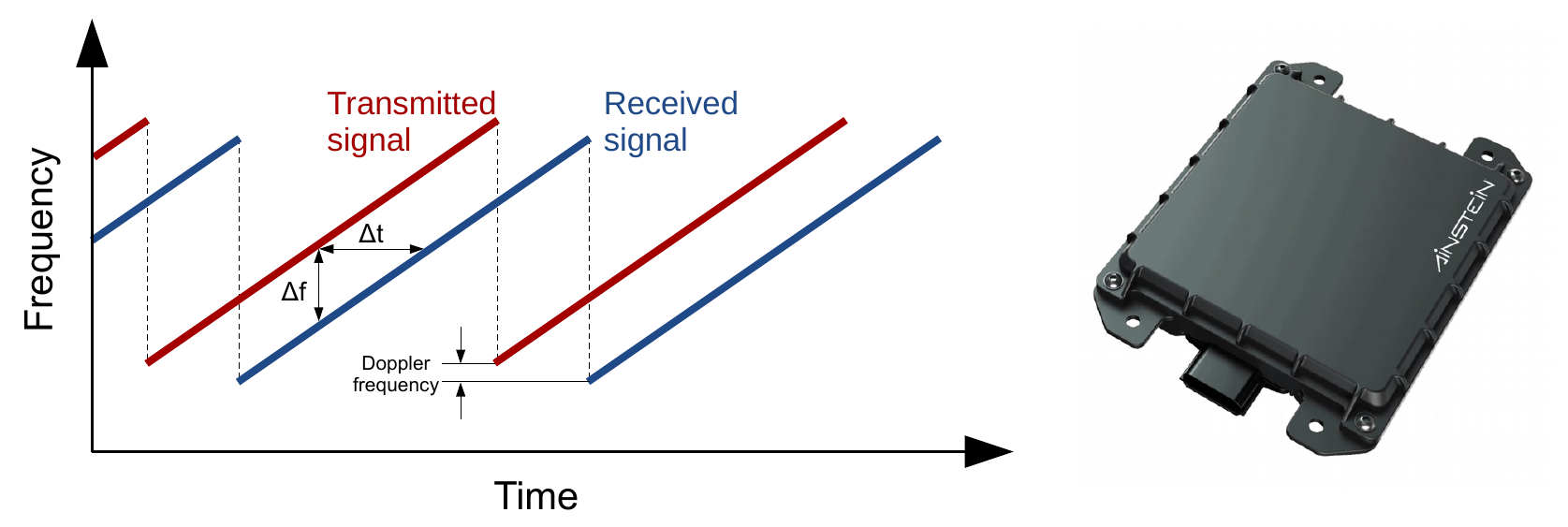}
   \caption{Operating principle of Frequency-Modulated Continuous-Wave (FMCW) radars (left); The Ainstein O-79 Imaging Radar (right).}
   \label{fig:radar}
\end{figure}

Similar to LiDARs, radar devices (acronym for Radio Detection And Ranging)
emit an electromagnetic wave, but uses radio waves (wavelengths typically between 3 mm and 30 cm) instead of laser light. The signal is transmitted and received by two distinct antennas or antenna arrays when required to estimate the angular positions of objects.
Compared to LiDARs, radars have lower angular and range resolution/accuracy but this disadvantage is compensated by the possibility of using
the Doppler effect to compute the relative velocity between the sensor and the detected objects. Another key benefit of radars is their robustness to harsh environmental conditions such as snow, fog, rain, dust etc.
Radars typically used in automotive and robotics are based on FMCW (Frequency-Modulated Continuous-Wave) transceivers, have a frequency range spanning from 20 to 80 GHz, and can detect the position, relative speed, and direction of motion of objects up to 200 meters.
FMCW radars emit a signal called \emph{chirp} where the frequency varies linearly over time. The difference between the frequency of the signal sent and that received is linearly related to the distance from the radar of the object that generated the reflected signal (see \figref{fig:radar}, left).
Nowadays, Cascade/Imaging Radar (CIR) systems are becoming very popular (e.g., \figref{fig:radar}, right).: in CIRs, multiple FMCW transceivers are cascaded together, to increase operational safety and achieve high angular resolution, thus enabling to generate a dense 3D point cloud mapping of the surrounding environment.

\subsection{Time of Flight Cameras}\label{sec:tof_cameras}

\begin{figure}[ht]
   \centering
   \includegraphics[width=\columnwidth]{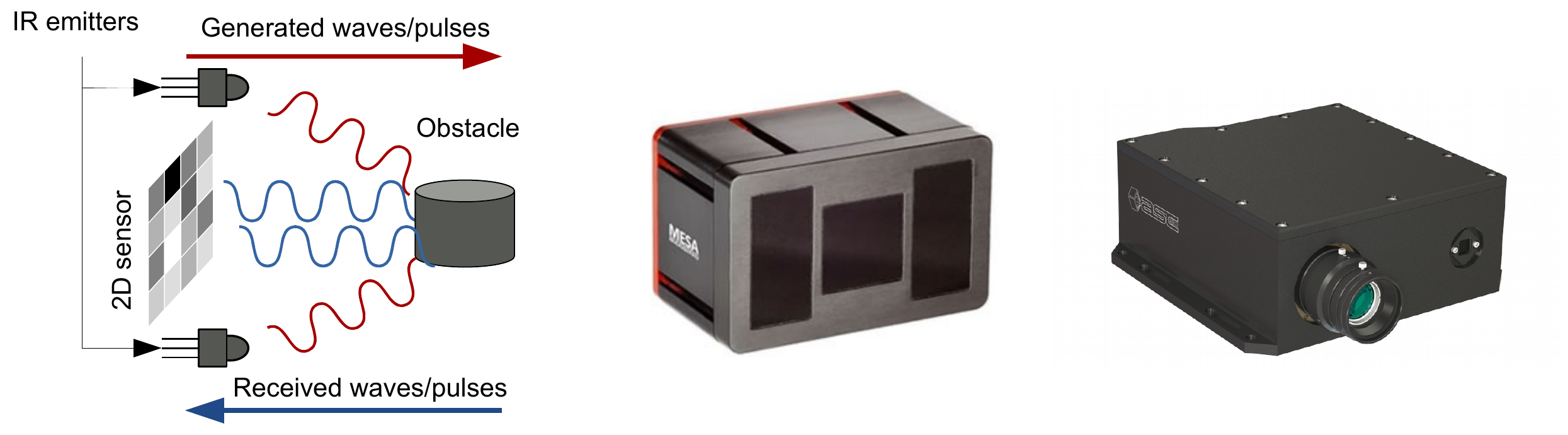}
   \caption{Operating principle of Tof cameras (left); MESA Imaging SR4500 continuous-wave ToF camera (center); Advanced Scientific Concepts GSFL-4K pulse-based ToF camera (right).}
   \label{fig:tof_cam}
\end{figure}

Time of Flight (ToF) Cameras can be considered as the meeting point between 
LiDARs and digital cameras. As LiDARs, they use projected laser light 
to measure the distance from the surrounding objects, following one of the
two main LiDAR's operating principles i.e., continuous-wave (e.g., 
\figref{fig:tof_cam}, center) or pulse-based (e.g.,  \figref{fig:tof_cam}, 
right), see \secref{sec:lidar}. As cameras, they employ a 
2D imaging sensor (e.g., a CCD or a CMOS) sensitive to the projected light.\\
ToF cameras typically use an array of infrared (IR) light emitters that illuminate the 
whole scene; the emitters are usually arranged around the IR 
imaging sensor, providing in output a depth map of the framed scene (\figref{fig:tof_cam}, left).
Differently from LiDARs, ToF cameras are able to measure distances of
a large portion of the scene in one single shot. On the other hand, the fact 
that they illuminate the whole scene and not individual points like LiDARs
introduces some important limitations, such as the \textit{multipath interference}
problem, caused by multiple projections of light rays coming 
from different objects into one single photoreceptor.
Since each light ray has a different phase shift, the sum of such components
will results in a wrong, random phase shift.

\subsection{Stereo Cameras}\label{sec:stereo_cameras}

\begin{figure}[h]
   \centering
   \includegraphics[width=\columnwidth]{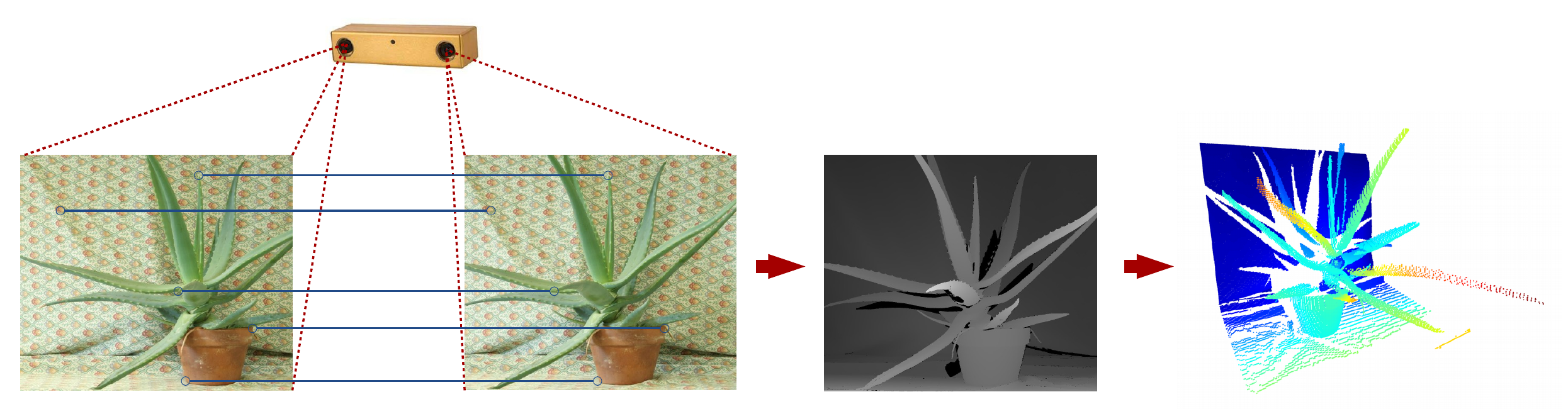}
   \caption{A Flir Bumblebee2 stereo camera framing a scene, with related \textit{left} and 
   \textit{right} images and some point correspondences 
   (left, images from the Middlebury Stereo Datasets); 
   the depth map estimated with dense stereo matching algorithms (center); a point cloud generated
   from the depth map (right). The point colors encode here the depth information 
   (red: closest points; blue: furthest points).}
   \label{fig:stereo}
\end{figure}

\textbf{Passive Stereo Cameras} are devices composed of two (or more) usually identical 
digital cameras, rigidly mounted on a common chassis and framing the same scene from different 
points of view. In most cases the cameras are mounted with a horizontal displacement.
The distance between the cameras is called \textit{baseline}: due to this non-zero baseline,
a 3D point is projected into the cameras' imaging sensors in different 2D points. 
Knowing the rigid body transformation that relates the cameras, from the coordinates of such 
2D points, it is possible to estimate the depth of the 3D point by using \textit{triangulation}
techniques. Exploiting this fact, the depth estimation problem can be implemented as an image-based
point correspondence problem: given an image point $p \in \mathbb{R}^2$ in a view (e.g., the left view), 
it is possible to estimate its depth by finding the point $p' \in \mathbb{R}^2$ in the other view 
that best matches $p$: $p$ and $p'$ should in fact represent projections of the same 3D point
$P \in \mathbb{R}^3$ (e.g., \figref{fig:stereo}, left). This problem is called 
\textit{stereo matching} and, if performed for each image point, \textit{dense}
stereo matching (e.g., \figref{fig:stereo}, center). 
The correspondence search is often speeded up by rectifying the images, which allows to
search for matches along image rows (\emph{scanlines}): in this case, $p$ and $p'$ belong
to the same row and differ by a displacement along this row called \textit{disparity}. 
The accuracy of stereo cameras decreases with the depth to be  measured.
Moreover, it is very hard to match points belonging to untextured, 
homogeneous areas: in this case, the depth is usually not estimated.\\

\begin{figure}[h]
\centering
\begin{minipage}[b]{0.5\columnwidth}
\includegraphics[width=\linewidth]{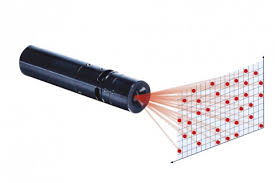}
\end{minipage}\hfill
\begin{minipage}[b]{0.25\columnwidth}
\includegraphics[width=\linewidth]{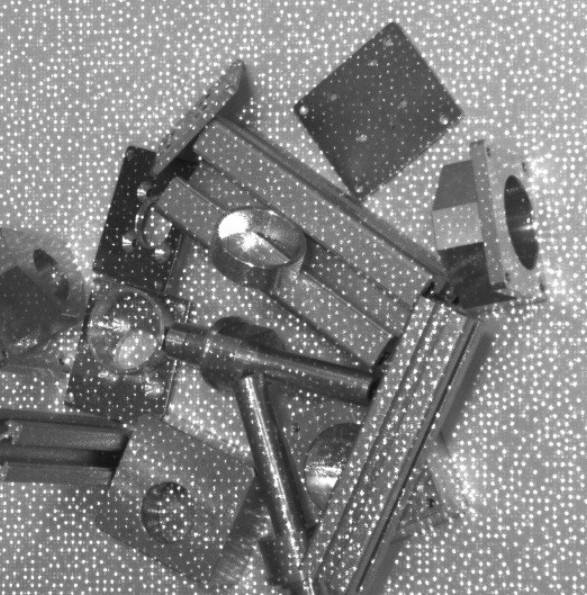}
\end{minipage}\hfill
\caption{The Osela Random Pattern Projector with an example of projected
divergent dot matrix pattern (left); 
an example of dot matrix pattern projected onto a scene (right).}   
\label{fig:projector}
\end{figure}

\textbf{Active Stereo Cameras} (e.g., \figref{fig:rgbd_camera}, left) solve this problem by coupling
to the digital cameras a dense pattern projector (e.g., \figref{fig:projector}) that projects into the scene
a visible textured pattern, so creating \textit{visual saliency} also in homogeneous surfaces.

\subsection{Structured Light Cameras}\label{sec:sl_cameras}

\begin{figure}[ht]
   \centering
   \includegraphics[width=\columnwidth]{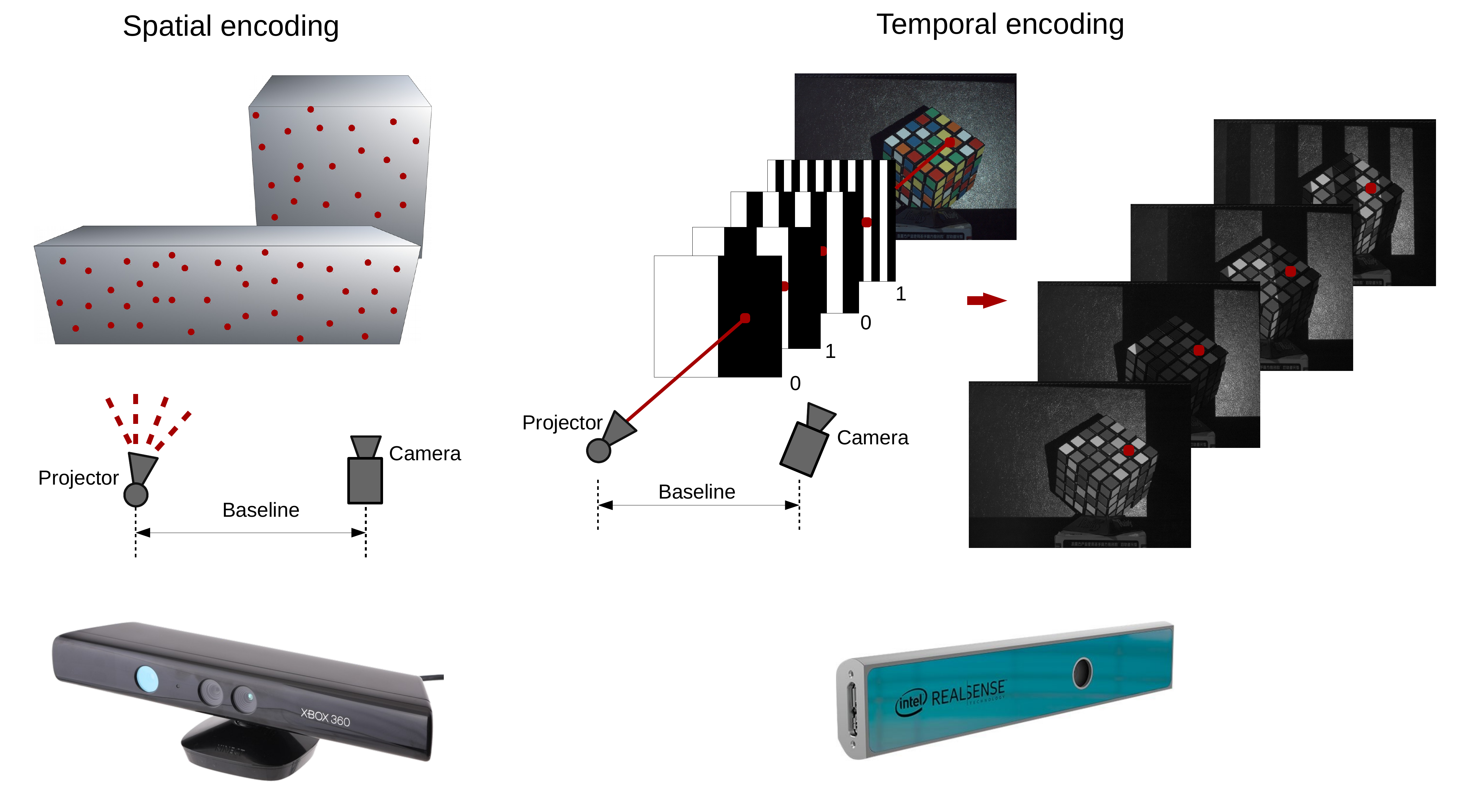}
   \caption{Operating principles (top row) of two types of SL cameras (bottom row): spatial encoding (red dots represent the projected pattern) used in the first version of the Microsoft Kinect (left column); temporal encoding used in the Intel RealSense SR305 (right column). Both the Kinect and SR305 also include an RGB camera to provide information on the color of the 3D points.}
   \label{fig:sl_camera}
\end{figure}

Structured Light (SL) cameras employ the same operating principle of stereo cameras
with a clear difference: one of the two cameras is replaced 
with a light projector that illuminates the scene with a textured visual pattern. 
The pattern projector can be seen as a virtual camera that 
always "sees" the same, fixed image: its projected pattern. The pattern is seen also 
by the camera but in this case, due to the baseline between the projector and the camera,
it is projected in different 2D points of the imaging sensor, depending on the 3D structure 
of the framed scene.
The projector can project a single pattern (spatial encoding) or a sequence of patterns (temporal encoding).\\

\textbf{Spatial encoding SL cameras} typically use visible or near-infrared dense pattern
projectors that illuminate the scene with a divergent dense pattern, commonly a
$n \times m$ pseudo random dot matrix pattern (e.g., see \figref{fig:projector}). 
The pattern is \textit{known} and each pattern patch (a $p \times p$ submatrix, $p \ll n,m$) 
represents a unique binary matrix. By detecting the pattern patches from the camera and
knowing the baseline between the camera and the pattern projector, it is 
possible to estimate the depths by means of triangulation (see \figref{fig:sl_camera}, left);\\

\textbf{Temporal encoding SL cameras} (also known as time-multiplexing SL cameras) project a fixed sequence of different patterns, capturing an image for each.
It is common to use binary patterns, each containing a sequence of vertical lines of equal width called \emph {fringes}, which either illuminate surface points with white light or not. The lines' width is halved for each consecutive pattern (e.g., see \figref{fig:sl_camera}, right).
If the system is rectified (see \secref{sec:stereo_cameras}) and $n$ binary patterns are projected, for each scanline it is possible to generate $2^n$ different binary codes. The camera-projector matching is directly recovered for each point from the binary code decoded by the time sequence of image intensities.\\

Generally, temporal encoding allows to obtain a better depth accuracy, and it is easier to implement. 
Spatial coding, on the other hand, allows to obtain depth maps with a single image, so 
it is more suitable for dynamic motions and/or dynamic environments. 

\subsection{RGB-D Cameras}\label{sec:rgbd_cameras}

\begin{figure}[h]
\centering
\begin{minipage}[b]{0.3\columnwidth}
\includegraphics[width=\linewidth]{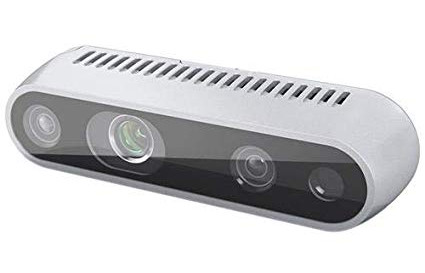}
\end{minipage}\hfill
\begin{minipage}[b]{0.3\columnwidth}
\includegraphics[width=\linewidth]{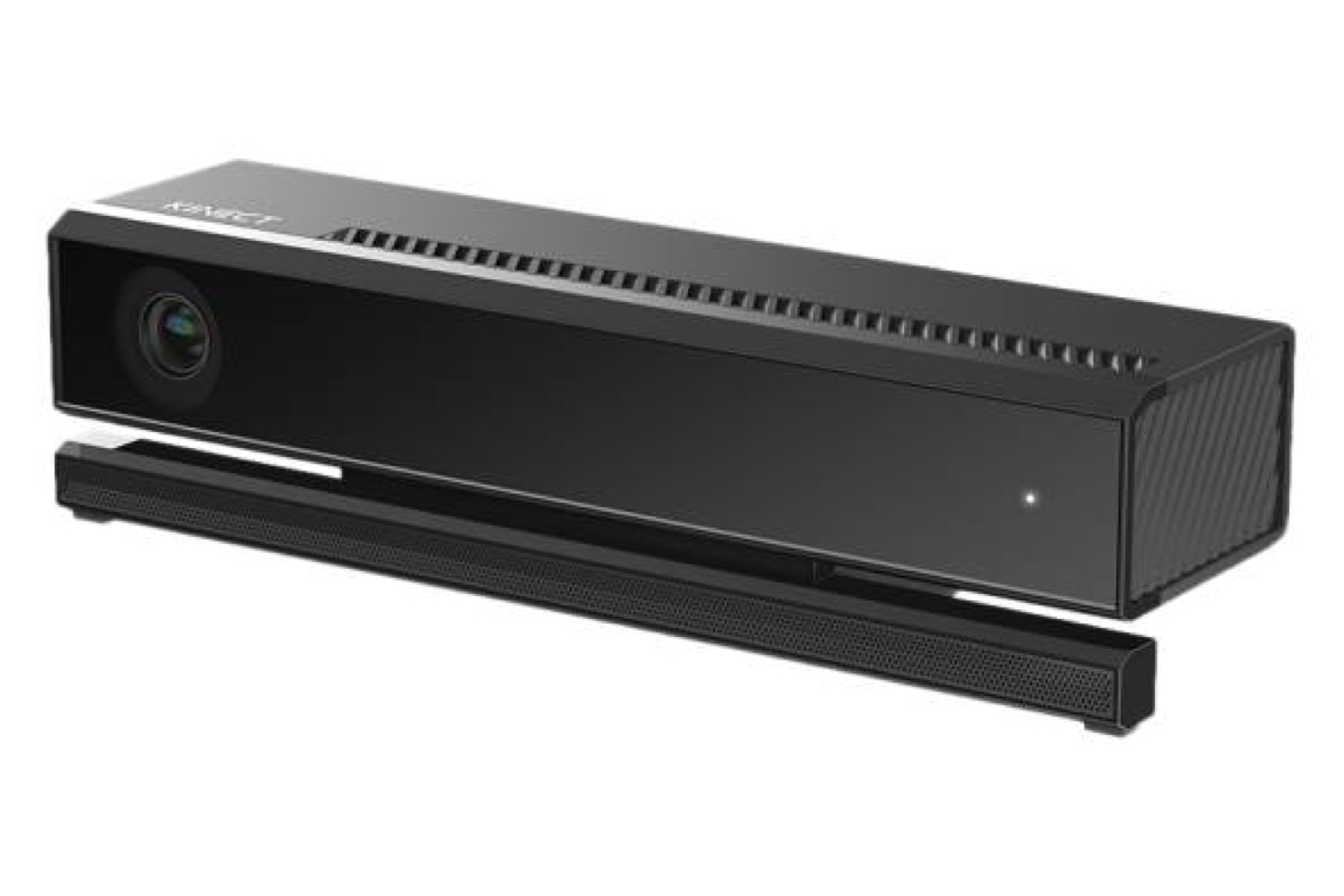}
\end{minipage}\hfill
\begin{minipage}[b]{0.3\columnwidth}
\includegraphics[width=\linewidth]{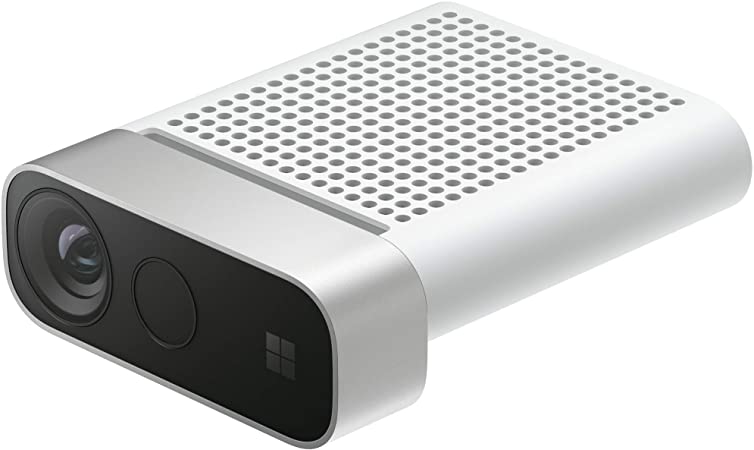}
\end{minipage}\hfill
\caption{Three example of RGB-D cameras: the Intel RealSense D435 (left); the Microsoft Kinect II (center); 
the Microsoft Azure Kinect (right). Both the Microsoft Kinect II and Azure Kinect use a CW ToF camera as ranging sensor,
while the D435 uses active stereo vision.}
\label{fig:rgbd_camera}
\end{figure}

RGB-D (RGB-Depth) cameras provide both color images and depth estimates.
Passive stereo cameras are natively RGB-D sensors, while other RGB-D sensors
are ensembles composed of a ToF, a SL, or an active stereo camera
rigidly coupled in the same chassis with a color camera
(e.g., \figref{fig:sl_camera} and \figref{fig:rgbd_camera}). 
The transformation that relates the two sensors is known, so it is possible
to acquire depth maps that are \textit{registered} with the related 
RGB images acquired by the color camera. In other words, each pixel of the RGB 
image is associated with a corresponding pixel in the depth map: both pixels
represent projections of the same 3D point into both sensors.
From this association is hence possible to generate a "colored" point cloud
that encodes both the structure and the visual appearance (i.e., the color)
of the framed scene.
This multimodal information enables one to tackle in a more effective way complex 
perception tasks such as recognizing and locating objects, 3D reconstruction of
environments, and detection and tracking of people. Recent RGB-D sensors as the
Microsoft Azure Kinect (\figref{fig:rgbd_camera}, right) also integrates microphone
arrays for speech and sound capture and IMUs for sensor orientation tracking.
%

\section{Example Applications}
To perform complex tasks and for safety reasons, mobile robots are often
equipped with a multitude of sensors. The choice of sensors depends on both
the application, the working environment, and the type of mobile robot.\\

A common feature of almost all mobile robots is the presence of encoders
(\secref{sec:encoders}) installed on the wheels. Wheel encoders are used to
estimate the robot \emph{odometry}, i.e. the position and velocity relative to a starting location. 
Such estimate is used in localization and SLAM problems as a motion prediction.\\

As introduced in \secref{sec:proximity}, contact and proximity sensors can
be used in any type of mobile robot as safety sensors to implement emergency
stop functions, or as simple reactive navigation sensors in limited indoor
environments (e.g., in robot vacuum cleaners). For more advanced safety functions,
for redundancy or for low-level obstacle avoidance functions both in indoor and
outdoor environments, rotating mirror LiDARs (\secref{sec:lidar}) or ultrasonic
sensor arrays (\secref{sec:sonar}) can also be used. The former, for example,
can be also certified for safety functions, while the latter can be used as
redundant sensors to allow detection of transparent surfaces.\\

LiDARs providing 2D scans are commonly used for indoor navigation (e.g., for 
localization and SLAM) often coupled with digital cameras (\secref{sec:cameras})
for higher-level tasks such as object detection and semantic segmentation
of scenes, with ToF cameras (\secref{sec:tof_cameras}) or SL cameras (\secref{sec:sl_cameras}) 
for 3D mapping and object pose estimation, or with RGB-D cameras (\secref{sec:rgbd_cameras})
for people detection and tracking and human–robot interaction. 
In GPS-denied environments such as large indoor industrial plants, 
AHRS IMUs (\secref{sec:imus}) can be used to provide a heading reference for navigation
purposes, while the environment can be structured with a UWB (\secref{sec:uwb}) network
to provide the robot with an absolute position reference.\\ 

In outdoor navigation, the use of a GPS (\secref{sec:gps}) receiver as an absolute position
reference is a common practice, while pulse-based spinning 3D LiDARs are often preferred as
range sensors, thanks to their wide field of view (typically 360-degree) and extended range
(up to 200 meters). In the case of harsh or critical outdoor
environments (for example, urban or agricultural environments) the LiDAR is often
coupled or replaced by one or more radars, that are more robust against dust,
rain or fog and, in case of quick motions, can provide velocity information
about other agents.
RGB cameras or passive stereo cameras are often used for outdoor 3D mapping,
place recognition, and loop closure detection and higher-level tasks such as
semantic segmentation, pedestrian detection, etc. RGB cameras, stereo cameras
and/or 3D LiDARS are also used, often coupled with IMUs, for robot ego-motion
estimation (\emph{visual odometry}), for instance, for redundancy or when
the odometry from wheel encoders is missing or unreliable due to slippery ground 
(e.g., when the robot moves off-road).

\section{Future Directions for Research}
In outlining the potential future of sensors for mobile robots, we first provide a short summary on comprehensive works on the state of the art. Based on these works and on our experience in the field, we then provide our view on the near future of these devices. Yet, we believe that doing forecasts in this domain is not an easy task, due to rapid technological development in fields such as semiconductors, material science, electronics, and manufacturing processes on which the construction of sensors relies. A breakthrough in one of these fields might lead to new classes of devices which we are unable to predict at the moment of writing.

A comprehensive and up-to-date review of the physical principles, design, and practical implementations of various transducers and sensors, with insights about the use of sensors
in mobile devices, can be found in \citep{Fraden2016}, while \citep{Everett1995} provides
in-depth analysis and details about many of the sensors specifically used in mobile robots.
A gentle yet exhaustive introduction to sensor's characterization and error modeling,
along with a detailed description of a large range of sensors for mobile robots, can be found in
the ''Perception`` chapter of \citep{Siegwart2nd2011}. \\
For most sensors, there are excellent specialized reference books and articles covering the theory of operation and advanced concepts, among others, for GPS/GNSS \citep{misra2011global}, MEMS IMUS \citep{kempe_2011}, digital cameras \citep{holst2011cmos}, omnidirectional cameras \citep{benosman2011panoramic}, event cameras \citep{9138762}, LiDARs \citep{shan2018topographic}, radars \citep{radar_principles_2010}, ToF cameras \citep{Horaud2016}, and SL cameras \citep{Zanuttigh2016}.\\

Very recently, a large number of new sensor technologies have appeared on the market, pushing the limits from a performance point of view and opening the door to new applications, often at affordable costs. These include OPA-based solid-state LiDARS \citep{Li2020-gm} (see also \secref{sec:lidar}), Multiple-Input, Multiple-Output (MIMO) radars \citep{9127853}, affordable  event cameras \citep{9138762},  low-cost multi-frequency multi-GNSS receivers \citep{NGUYEN2021100004}, smart AI (Artificial Intelligent) devices embedding microcontrollers able to locally execute deep neural network learning workloads \citep{s21134412}, and others. In the next few years, 360‑degree solid-state scanning LiDAR \citep{Nishiwaki2021-fg} and low-cost megapixel-resolution depth cameras based on piezoelectric effect \citep{Atalar2022-nb} could likely enter the market, making the spatial perception of mobile robots even more effective.\\

On the other hand, there is still a lot of room for research in the sensors field, especially in the case of mobile robots, which often include heterogeneous sensor ensembles and require high-level information to autonomously and effectively interact with the environment. For example, sensors' ability to self-calibrate and self-configure regardless of their arrangement in the robot could be a crucial enabling technology. Smart sensors that, thanks to embedded AI models, directly output high-level information (e.g., object classes with poses and dynamics) could speed up the implementation of new, cutting-edge applications.
\bibliographystyle{spbasic}
\bibliography{s4mr_ref}

\end{document}